\pdfoutput=1

\documentclass[11pt]{article}

\usepackage[]{acl}

\usepackage{times}
\usepackage{latexsym}

\usepackage[T1]{fontenc}

\usepackage[utf8]{inputenc}

\usepackage{adjustbox}
\usepackage{multirow}

\usepackage{microtype}

\usepackage{xspace}
\usepackage{mathtools}
\usepackage{amssymb}
\usepackage{rotating}
\usepackage{footnote}
\usepackage{tablefootnote}
\usepackage{graphicx}
\usepackage{color, colortbl}
\usepackage{xstring}
\usepackage{comment}
\usepackage{caption}
\usepackage{subcaption}
\usepackage{float}
\restylefloat{table}
\usepackage{array,booktabs,makecell}
\usepackage{appendix}
\usepackage[normalem]{ulem}

\usepackage{enumitem}

\usepackage{scalerel,xparse}
\usepackage{siunitx}

\newcommand{\eat}[1]{}

 \newcommand\Tau{\scalerel*{\tau}{T}}

\newcolumntype{H}{>{\setbox0=\hbox\bgroup}c<{\egroup}@{}}

\usepackage{arydshln}

\PassOptionsToPackage{hyphens}{url}\usepackage{hyperref}
  
\NewDocumentCommand\emojisick{}{
    \scalerel*{
        \includegraphics{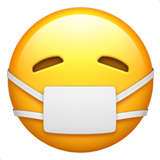}
    }{X}
}

\NewDocumentCommand\emojidispoint{}{
    \scalerel*{
        \includegraphics{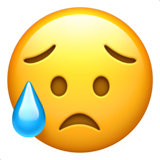}
    }{X}
}

\NewDocumentCommand\emojianger{}{
    \scalerel*{
        \includegraphics{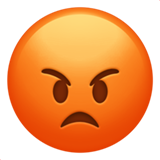}
    }{X}
}

\NewDocumentCommand\emojiupsidedown{}{
    \scalerel*{
        \includegraphics{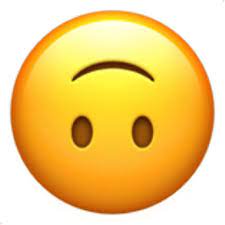}
    }{X}
}

\title{Improving Social Meaning Detection with Pragmatic Masking and Surrogate Fine-Tuning}

\author{Chiyu Zhang  ~~~~Muhammad Abdul-Mageed  \\ 
  Deep Learning \& Natural Language Processing Group \\The University of British Columbia \\
  \tt chiyuzh@mail.ubc.ca, \tt muhammad.mageed@ubc.ca}

\begin{document}
\maketitle
\begin{abstract}
Masked language models (MLMs) are pre-trained with a denoising objective that is in a mismatch with the objective of downstream fine-tuning. We propose pragmatic masking and surrogate fine-tuning as two complementing strategies that exploit social cues to drive pre-trained representations toward a broad set of concepts useful for a wide class of social meaning tasks.   
We test our models on $15$ different Twitter datasets for social meaning detection. Our methods achieve $2.34\%$ \textit{F}$_1$ over a competitive baseline, while outperforming domain-specific language models pre-trained on large datasets. Our methods also excel in few-shot learning: with only $5\%$ of training data (severely few-shot), our methods enable an impressive $68.54\%$ average \textit{F}$_1$. The methods are also language agnostic, as we show in a zero-shot setting involving six datasets from three different languages.\footnote{Our code is available at: \url{https://github.com/chiyuzhang94/PMLM-SFT}.}
\end{abstract}

\section{Introduction}\label{sec:intro}
Masked language models (MLMs) such as BERT~\cite{devlin-2019-bert} have revolutionized natural language processing (NLP). These models exploit the idea of self-supervision where sequences of unlabeled text are masked and the model is tasked to reconstruct them. Knowledge acquired during this stage of denoising (called \textit{pre-training}) can then be transferred to downstream tasks through a second stage (called \textit{fine-tuning}). Although pre-training is general, does not require labeled data, and is task agnostic, fine-tuning is narrow, requires labeled data, and is task-specific. For a class of tasks $\Tau$, some of which we may not know in the present but which can become desirable in the future, it is unclear how we can bridge the learning objective mismatch between these two stages. In particular, how can we \textbf{(i)} make pre-training more tightly related to downstream task learning objective; and \textbf{(ii)} focus model pre-training representation on an all-encompassing range of concepts of general affinity to various downstream tasks?

\begin{figure}[t]
\centering
\begin{subfigure}[]{\linewidth}
  \centering
  \includegraphics[width=\linewidth]{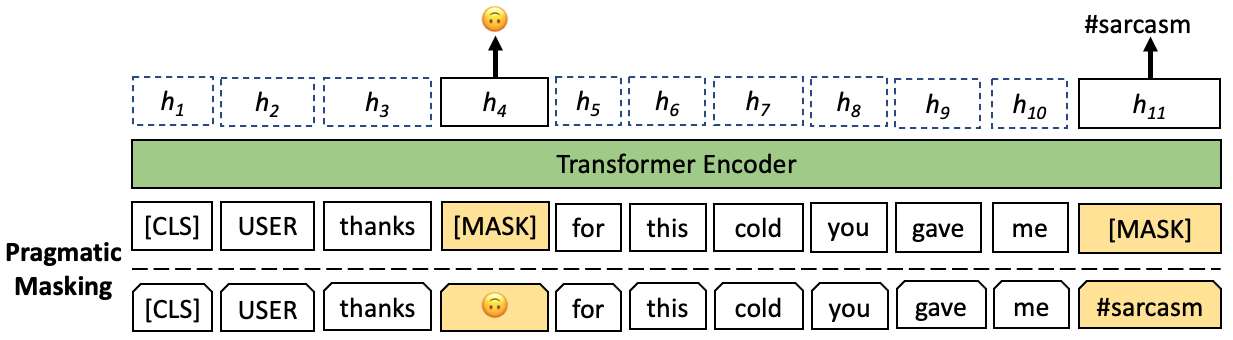}
  \caption{Pragmatic masking}\label{fig:pm}
\end{subfigure} 
\begin{subfigure}[]{\linewidth}
  \centering
  \includegraphics[width=\linewidth]{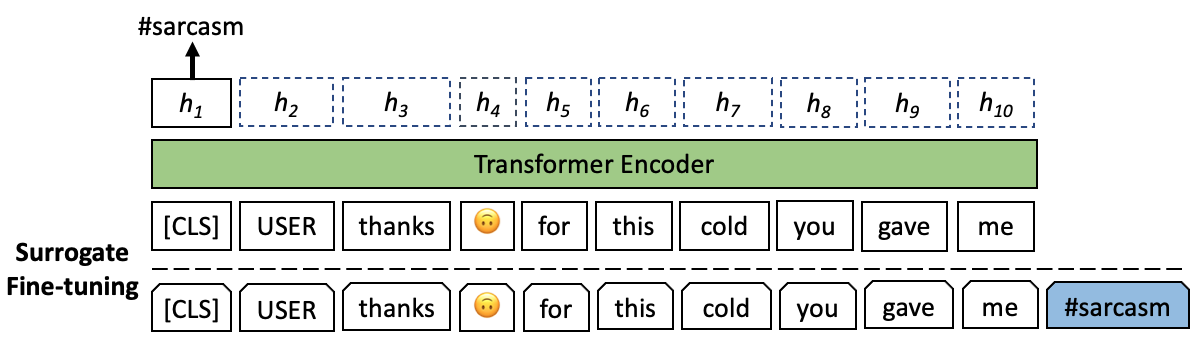}
  \caption{Surrogate fine-tuning}\label{fig:sft}
\end{subfigure}
\caption{Illustration of our proposed pragmatic masking and surrogate fine-tuning methods.}
\end{figure}

We raise these questions in the context of learning a cluster of tasks to which we collectively refer as \textit{social meaning}. We loosely define social meaning as meaning emerging through human interaction such as on social media. Example social meaning tasks include emotion, irony, and sentiment detection. We propose two main solutions that we hypothesize can bring pre-training and fine-tuning closer in the context of learning social meaning: First, we propose a particular type of guided masking that prioritizes learning contexts of tokens 
crucially relevant to social meaning in interactive discourse. Since the type of ``meaning in interaction'' we are interested in is the domain of linguistic pragmatics~\cite{thomas2014meaning}, we will refer to our proposed masking mechanism as \textit{pragmatic masking}. We explain pragmatic masking in Section~\ref{sec:prag_masking}. 

Second, we propose an additional novel stage of fine-tuning that does not depend on gold labels but instead exploits general data cues possibly relevant to \textit{all} social meaning tasks. More precisely, we leverage sequence-level user assigned tags for \textit{intermediate} fine-tuning of pre-trained language models. In the case of Twitter, for example, hashtags naturally assigned by users at the end of posts can carry discriminative power that is by and large relevant to a wide host of tasks. Although cues such as hashtags and emojis have been previously used as surrogate lables before for one task or another, we put them to a broader use that is not focused on a particular (usually narrow) task that learns from a handful of cues. In other words, our goal is to learn extensive concepts carried by tens of thousands of cues. A model endowed with such a knowledge-base of social concepts can then be further fine-tuned on any narrower task in the ordinary way. We refer to this method as \textit{surrogate fine-tuning} (Section~\ref{sec:surrogate_ft}). Another migration from previous work is that our methods excel not only in the full-data setting but also for \textit{few-shot learning}, as we will explain below.
In order to evaluate our methods, we present a social meaning benchmark composed of $15$ different datasets crawled from previous research sources. We perform an extensive series of methodical experiments directly targeting our proposed methods. Our experiments set new state-of-the-art (SOTA) in the supervised setting across different datasets. Moreover, our experiments reveal a striking capacity of our models in improving downstream task performance in few-shot and severely few-shot settings (i.e., as low as $1\%$ of gold data), and even the zero-shot setting on languages other than English (i.e., as evaluated on six different datasets from three languages in Section~\ref{sec:zfew}). 

To summarize, we make the following \textbf{contributions}:
 \textbf{(1)} We propose a novel pragmatic masking strategy that makes use of social media cues akin to improving social meaning detection. \textbf{(2)} We introduce a new effective surrogate fine-tuning method suited to social meaning that exploits the same simple cues as our pragmatic masking strategy. \textbf{(3)} We report new SOTA on eight out of $15$ supervised datasets in the full-data setting. \textbf{(4)} Our methods are remarkably effective for few-shot and zero- and learning. We now review related work.
 


\vspace{-3pt}
\section{Related works}\label{sec:related_work}
\vspace{-3pt}
\textbf{Masked Language models.} 
\citet{devlin-2019-bert} introduced BERT, a language representation model pre-trained by joint conditioning on both left and right context in all layers with the Transformer encoder~\cite{vaswani2017attention}. BERT's pre-training introduces a self-supervised learning objective, i.e., masked language modeling (MLM), to train the Transformer encoder. MLM predicts masked tokens in input sequences exploiting bi-directional context. 
RoBERTa~\cite{liu2019roberta} optimizes BERT performance by removing the next sentence prediction objective and by pre-training on a larger corpus using a bigger batch size. In the last few years, several variants of LMs with different masking methods were proposed. Examples are XLNet~\cite{yang2019xlnet} and MASS~\cite{song2019mass}. 
To incorporate more domain specific knowledge into LMs, some works introduce knowledge-enabled masking strategies. For example, \citet{sun2019ernie,zhang-2019-ernie,lin-2021-entitybert} propose to mask tokens of named entities, while 
\citet{tian-2020-skep} and \citet{ke-2020-sentilare} select sentiment-related words to mask during pre-training. \citet{gu-2020-train} and \citet{kawintiranon-2021-knowledge} propose selective masking methods to mask the more important tokens for downstream tasks (e.g., sentiment analysis and stance detection). However, these masking strategies depend on external resources and/or annotations (e.g., a lexicon or labeled corpora).~\citet{corazza-2020-hybrid} investigate the utility of hybrid emoji-based masking for enhancing abusive language detection. Previous works, therefore, only focus on one or another particular task (e.g., sentiment, abusive language detection) rather than the type of broad representations we target.

\textbf{Intermediate Fine-Tuning.} Although pre-trained language models (PLM) have shown significant improvements on NLP tasks, intermediate training of the PLM on one or more data-rich tasks can further improve performance on a target downstream task. Most previous work (e.g., ~\cite{wang-2019-tell,pruksachatkun-2020-intermediate,phang-2020-english,chang-2021-rethinking,poth-2021-pre}) focus on intermediate fine-tuning on a given gold-labeled dataset related to a downstream target task. Different to these works, our surrogate fine-tuning method is \textit{agnostic} to narrow downstream tasks and fine-tunes an PLM on large-scale data with tens of thousands of \textit{surrogate} labels that may be relevant to all social meaning. We now introduce our methods. 

\vspace{-5pt}
\section{Proposed Methods}\label{sec:method}
\vspace{-3pt}

\subsection{Pragmatic Masking}\label{sec:prag_masking} 
MLMs employ random masking, and so are not guided to learn any particular type of information during pre-training. Several attempts have been made to employ task-specific masking where the objective is to predict information relevant to a given downstream task. Task relevant information is usually identified based on world knowledge (e.g., a sentiment lexicon~\cite{gu-2020-train,ke-2020-sentilare}, part-of-speech (POS) tags~\cite{zhou-2020-limit}) or based on some other type of modeling such as pointwise mutual information~\cite{tian-2020-skep} with supervised data. Although task-specific masking is useful, it is desirable to identify a \textit{more general} masking strategy that \textit{does not depend on external information} that may not be available or hard to acquire (e.g., costly annotation). For example, there are no POS taggers for some languages and so methods based on POS tags would not be applicable. Motivated by the fact that random masking is intrinsically sub-optimal~\cite{ke-2020-sentilare,kawintiranon-2021-knowledge} and this particular need for a more general and dependency-free masking method, we introduce our novel pragmatic masking mechanism that is suited to a wide range of social meaning tasks. 
\begin{table}[t]
\small
\centering
\renewcommand{\arraystretch}{1.5}
\begin{tabular}{l}
\toprule
\begin{tabular}[c]{@{}l@{}}\textbf{(1)} Just got chased through my house with a bowl of tuna\\  fish.\emojisick ing.\hfill \hfill \hfill \hfill \hfill\textbf{[Disgust]} \end{tabular}                                                                                                                                \\ \cdashline{1-1} 
\begin{tabular}[c]{@{}l@{}}\textbf{(2)} USER thanks \emojiupsidedown for this cold you gave me \#sarcasm  \\ \hfill \hfill \hfill \hfill \hfill \hfill \hfill \hfill \hfill \hfill \hfill \hfill \hfill \textbf{[Sarcastic]} \end{tabular}                                                                                            \\ \cdashline{1-1} 
\begin{tabular}[c]{@{}l@{}}\textbf{(3)} USER Awww \emojidispoint\emojidispoint CUPCAKES SUCK IT UP. SHE \\ LOST \emojianger\emojianger GET OVER IT \emojianger\emojianger \hfill \hfill   \textbf{[Offensive]}\end{tabular} \\ \toprule
\end{tabular}
\caption{Samples from our social meaning benchmark.}\label{tab:tw_sample}
\end{table}

To illustrate, consider the tweet samples in Table~\ref{tab:tw_sample}: In example (1), the emoji ``\emojisick'' combined with the suffix ``-ing'' in ``\emojisick ing'' is a clear signal indicating the \textit{disgust} emotion. In example (2) the emoji ``\emojiupsidedown'' and the hashtag ``\#sarcasm'' communicate \textit{sarcasm}. In example (3) the combination of the emojis ``\emojidispoint'' and ``\emojianger'' accompany `hard' emotions characteristic of \textit{offensive} language. We hypothesize that by simply masking cues such as emojis and hashtags, we can bias the model to learn about different shades of social meaning expression. This masking method can be performed in a \textit{self-supervised} fashion since hashtags and emojis can be automatically identified. We call the resulting language model \textbf{\textit{pragmatically masked language model (PMLM)}}. Specifically, when we choose tokens for masking, we prioritize hashtags and emojis as Figure~\ref{fig:pm} shows.
The pragmatic masking strategy follows several steps:
\noindent \textbf{(1) Pragmatic token selection.} 
We randomly select up to $15\%$ of input sequence, giving masking \textbf{priority} to hashtags or emojis. The tokens are selected by whole word masking (i.e., whole hashtag or emoji). \noindent \textbf{(2) Regular token selection.} If the pragmatic tokens are less than $15\%$, we then randomly select regular BPE tokens to complete the percentage of masking to the $15\%$. \noindent \textbf{(3) Masking.} This is the same as the RoBERTa MLM objective where we replace $80\%$ of selected tokens with the [MASK] token, $10\%$ with random tokens, and we keep $10\%$ unchanged.

\subsection{Surrogate Fine-tuning}\label{sec:surrogate_ft}
The current transfer learning paradigm of first pre-training then fine-tuning on particular tasks is limited by how much labeled data is available for downstream tasks. In other words, this existing set up works only given large amounts of labeled data. We propose surrogate fine-tuning where we intermediate fine-tune PLMs to predict thousands of example-level cues (i.e., hashtags occurring at the end of tweets) as Figure~\ref{fig:sft} shows. This method is inspired by previous work that exploited hashtags~\cite{riloff2013sarcasm,ptavcek2014sarcasm,rajadesingan2015sarcasm,sintsova-2016-dystemo,abdul-2017-emonet,barbieri-2018-semeval} or emojis~\cite{wood2016emoji,felbo-2017-using,wiegand-2021-exploiting} as proxy for labels in a number of social meaning tasks. 
However, instead of identifying a small \textit{specific} set of hashtags or emojis for a \textit{single} task and using them to collect a dataset of \textit{distant} labels, we diverge from the literature in proposing to use data with \textit{any} hashtag or emoji as a surrogate labeling approach suited for \textit{any} (or at least most) social meaning task. 
As explained, we refer to our method as \textbf{\textit{surrogate fine-tuning (SFT).}}

\section{Experiments}\label{sec:experiment}

\subsection{Pre-training Data}
\noindent \textbf{TweetEnglish Dataset.}
We extract $2.4$B English tweets\footnote{We select English tweets based on the Twitter language tag.} from a larger in-house dataset collected between $2014$ and $2020$. We lightly normalize tweets by removing usernames and hyperlinks and add white space between emojis to help our model identify individual emojis. We keep all the tweets, retweets, and replies but remove the `RT USER:' string in front of retweets. To ensure each tweet contains sufficient context for modeling, we filter out tweets shorter than $5$ English words (not counting the special tokens hashtag, emoji, USER, URL, and RT). We call this dataset \textbf{\texttt{TweetEng}}. Exploring the distribution of hashtags and emojis within TweetEng, we find that $18.5$\% of the tweets include at least one hashtag but no emoji, $11.5$\% have at least one emoji but no hashtag, and $2.2$\% have both at least one hashtag and at least one emoji. Investigating the hashtag and emoji location, we observe that $7.1$\% of the tweets use a hashtag as the last term, and that the last term of $6.7$\% of tweets is an emoji. We will use \texttt{TweetEng} as a general pool of data from which we derive for both our PMLM and SFT methods. 

\textbf{PM Datasets.}\label{subsec:prag_masking:data}
We extract five different subsets from \texttt{TweetEng} to explore the utility of our proposed PMLM method. Each of these five datasets comprises $150$M tweets as follows:
\texttt{\textbf{Naive}}. A randomly selected tweet set. Based on the distribution of hashtags and emojis in \texttt{TweetEng}, each sample in \texttt{Naive} still has some likelihood to include one or more hashtags and/or emojis. We are thus still able to perform our PM method on \texttt{Naive}. \texttt{\textbf{Naive-Remove}}. To isolate the utility of using pragmatic cues, we construct a dataset by removing all hashtags and emojis from \texttt{Naive}. \texttt{\textbf{Hashtag\_any}}. Tweets with at least one hashtag anywhere but no emojis. \texttt{\textbf{Emoji\_any}}. Tweets with at least one emoji anywhere but no hashtags. \texttt{\textbf{Hashtag\_end}}. Tweets with a hashtag as the last term but no emojis. \texttt{\textbf{Emoji\_end}}. Tweets with an emoji at the end of the tweet but no hashtags.\footnote{We perform an analysis based on two 10M random samples of tweets from Hashtag\_any and Emoji\_any, respectively. We find that on average there are 1.83 hashtags per tweet in Hashtag\_any and 1.88 emojis per tweet in Emoji\_any.}

\noindent\textbf{SFT Datasets.}\label{subsec:sft:data}
We experiment with two SFT settings, one based on \textit{hashtags} (\textit{SFT-H}) and another based on \textit{emojis} (\textit{SFT-E}). For SFT-H, we utilize the \texttt{Hashtag\_end} dataset mentioned above. The dataset includes $5$M unique hashtags (all occurring at the end of tweets), but the majority of these are low frequency. We remove any hashtags occurring $< 200$ times, which gives us a set of $63K$ hashtags in $126M$ tweets. We split the tweets into Train ($80\%$), Dev ($10\%$), and Test ($10\%$). For each sample, we use the end hashtag as the sample label.\footnote{We use the last hashtag as the label if there are more than one hashtag in the end of a tweet. Different from PMLM, SFT is a multi-class single-label classification task. We plan to explore the multi-class multi-label SFT in the future.} We refer to this resulting dataset as \textbf{\texttt{Hashtag\_pred}}. For emoji SFT, we work with the \texttt{emoji\_end} dataset. Similar to SFT-H, we remove low-frequence emojis ($< 200$ times), extract the same number of tweets as \texttt{Hashtag\_pred}, and follow the same data splitting method. We acquire a total of $1,650$ unique emojis in final positions, which we assign as class labels and remove them from the original tweet body. We refer to this dataset as \textbf{\texttt{Emoji\_pred}}.

\subsection{Evaluation Benchmark}\label{sec:smpb} 
We collect $15$ datasets representing eight different social meaning tasks to evaluate our models, as follows:~\footnote{To facilitate reference, we give each dataset a name.}

\noindent\textbf{Crisis awareness.}~We use \texttt{Crisis\textsubscript{Oltea}} ~\cite{olteanu2014crisislex}, a corpus for identifying whether a tweet is related to a given disaster or not. 
    
 \noindent     \textbf{Emotion.} We utilize \texttt{Emo\textsubscript{Moham}}, introduced by~\citet{mohammad-2018-semeval}, for emotion recognition. We use the version adapted in \citet{barbieri-2020-tweeteval}.
 
\noindent\textbf{Hateful and offensive language.} We use \texttt{Hate\textsubscript{Waseem}}~\cite{waseem-2016-hateful}, \texttt{Hate\textsubscript{David}}~\cite{davidson-2017-hateoffensive}, and \texttt{Offense\textsubscript{Zamp}}~\cite{zampieri-2019-predicting}.

\noindent     \textbf{Humor.} We use the humor detection datasets \texttt{Humor\textsubscript{Potash}}~\cite{potash-2017-semeval} and \texttt{Humor\textsubscript{Meaney}}~\cite{meaney2021hahackathon}.

\noindent\textbf{Irony.} We utilize \texttt{Irony\textsubscript{Hee-A}} and  \texttt{Irony\textsubscript{Hee-B}} from \citet{van-hee2018semeval}. 
  
\noindent     \textbf{Sarcasm.} 
We use four sarcasm datasets from \texttt{Sarc\textsubscript{Riloff}}~\cite{riloff2013sarcasm}, \texttt{Sarc\textsubscript{Ptacek}}~\cite{ptavcek2014sarcasm}, \texttt{Sarc\textsubscript{Rajad}}~\cite{rajadesingan2015sarcasm}, and \texttt{Sarc\textsubscript{Bam}}~\cite{bamman2015contextualized}. 
 
\noindent\textbf{Sentiment.} We employ the three-way sentiment analysis dataset from \texttt{Senti\textsubscript{Rosen}}~\cite{rosenthal-2017-semeval}. 

\noindent\textbf{Stance.} We use \texttt{Stance\textsubscript{Moham}}, a stance detection dataset from \citet{mohammad-2016-semeval}. The task is to identify the position of a given tweet towards a target of interest. 

We use the Twitter API~\footnote{\url{https://developer.twitter.com/}} to crawl datasets which are available only in tweet ID form. We note that we could not download all tweets since some tweets get deleted by users or become inaccessible for some other reason. Since some datasets are old (dating back to 2013), we are only able to retrieve $73\%$ of the tweets on average (i.e., across the different datasets). 
We normalize each tweet by replacing the user names and hyperlinks to the special tokens `USER' and `URL', respectively. 
For datasets collected based on hashtags by original authors (i.e., \textit{distant supervision}), we also remove the seed hashtags from the original tweets. For datasets originally used in cross-validation, we acquire $80\%$ Train, $10\%$ Dev, and $10\%$ Test via random splits. For datasets that had training and test splits but not development data, we split off $10\%$ from training data into Dev. The data splits of each dataset are presented in Table~\ref{tab:gold_data}. 
\begin{table}[h]
\centering
\tiny
{%
\begin{tabular}{@{}lclrrrH@{}}
\toprule
\multicolumn{1}{c}{\textbf{Task}}         & \multicolumn{1}{c}{\textbf{Lg}}       & \multicolumn{1}{l}{\textbf{Classes}}                            & \multicolumn{1}{c}{\textbf{Train}} & \multicolumn{1}{c}{\textbf{Dev}} & \multicolumn{1}{c}{\textbf{Test}} & \multicolumn{1}{H}{\textbf{Total}} \\ \midrule
Crisis\textsubscript{Oltea}
& EN & \{on-topic, off-topic,\}                                        & $48.0$K                             & $6.0$K                            & $6.0$K                             & $60.0$K                             \\
Emo\textsubscript{Moham}
& EN & \{anger, joy, opt., sad.\}                               & $3.3$K                              & $374$                              & $1.4$K                             & $5.0$K                              \\

Hate\textsubscript{Waseem}
& EN  & \{racism, sexism, none\}                                        & $8.7$K                              & $1.1$K                            & $1.1$K                             & $10.9$K                             \\
Hate\textsubscript{David}
& EN & \{hate, off., neither\}                                    & $19.8$K                             & $2.5$K                            & $2.5$K                            & $24.8$K                            \\
Humor\textsubscript{Potash}
& EN & \{humor, not humor\}                          & $11.3$K                             & $660$                              & $749$                               & $12.7$K                             \\
Humor\textsubscript{Meaney}
& EN &                     \{humor, not humor\}                                             & $8.0$K                             & $1.0$K                              & $1.0$K                             & $10.0$K                             \\
Irony\textsubscript{Hee-A}
& EN & \{ironic, not ironic\}                                          & $3.5$K                             & $384$                              & $784$                               & $4.6$K                             \\
Irony\textsubscript{Hee-B}
& EN & \{IC, SI, OI, NI\} & $3.5$K                              & $384$                              & $784$                               & $4.6$K                              \\
Offense\textsubscript{Zamp}
& EN & \{off., not off.\}                                   & $11.9$K                             & $1.3$K                           & $860$                               & $14.1$K                            \\
Sarc\textsubscript{Riloff}
& EN & \{sarc., non-sarc.\}                   & $1.4$K                             & $177$                              & $177$                               &$1.8$K                             \\
Sarc\textsubscript{Ptacek}
& EN &    \{sarc., non-sarc.\}                                                              & $71.4$K                             & $8.9$K                            & $8.9$K                             & $89.3$K                          \\
Sarc\textsubscript{Rajad}
& EN &   \{sarc., non-sarc.\}                                                               & $41.3$K                            & $5.2$K                          & $5.2$K                           & $51.6$K                          \\
Sarc\textsubscript{Bam}
& EN &          \{sarc., non-sarc.\}                                                        & $11.9$K                             & $1.5$K                          & $1.5$K                             & $14.8$K                             \\
Senti\textsubscript{Rosen}
& EN & \{neg., neu., pos.\}                                 & $42.8$K                            & $4.8$K                           & $12.3$K                         & $59.8$K                           \\
Stance\textsubscript{Moham}
& EN & \{against, favor, none\}                                        & $2.6$K                              & $292$                              & $1.3$K                             & $4.2$K                             \\ \cdashline{1-7}
Emo\textsubscript{Mageed}
& AR & \{anger, joy, sad.\}                                        & -                              & -                              & $372$                             & -                             \\
Irony\textsubscript{Ghan}
& AR & \{ironic, not ironic\}                                        & -                              & -                              & $805$                             & -                             \\
Emo\textsubscript{Bian}
& IT & \{anger, joy, sad.\}                                        & -                              & -                              & $196$                             & -                             \\
Hate\textsubscript{Bosco}
& IT & \{hate, not hate\}                                        & -                              & -                              & $1.0$K                             & -                            \\
Emo\textsubscript{Moham}
& ES & \{anger, joy, sad.\}                                        & -                              & -                             & $2.0$K                             & -                             \\
Hate\textsubscript{Bas}
& ES & \{hate, not hate\}                                        & -                              & -                             & $1.6$K                             & -                             \\\bottomrule
\end{tabular}%
}
\caption{Social meaning data. \textbf{opt.:}: Optimism, \textbf{sad.:} Sadness, \textbf{off.:} offensive, \textbf{sarc.:} sarcastic, \textbf{IC:} Ironic by clash, \textbf{SI:} Situational irony, \textbf{OI:} Other irony, \textbf{NI:} Non-ironic, \textbf{neg.:} Negative, \textbf{neu.:} Neutral, \textbf{pos.:} Positive.}\label{tab:gold_data}
\end{table}

To test our models under the \textbf{\textit{few-shot setting}}, we conduct few-shot experiments on varying percentages of the Train set of each task (i.e., $1\%$, $5\%$, $10\%$, $20\%$ \dots $90\%$). For each of these sizes, we randomly sample three times with replacement (\textit{as we report the average of three runs in our experiments}) and evaluate each model on the original Dev and Test sets. We also evaluate our models on the \textbf{\textit{zero-shot setting}} utilizing data from Arabic: \texttt{Emo\textsubscript{Mageed}}~\cite{mageed-2020-aranet}, \texttt{Irony\textsubscript{Ghan}}~\cite{idat2019}; Italian: \texttt{Emo\textsubscript{Bian}}~\cite{bianchi2021feel} and \texttt{Hate\textsubscript{Bosco}}~\cite{bosco2018overview}; and Spanish:    \texttt{Emo\textsubscript{Moham}}~\cite{mohammad-2018-semeval} and \texttt{Hate\textsubscript{Bas}}~\cite{basile-2019-semeval}. 

 

\subsection{Implementation and Baselines}\label{subsec:baseline} 
For both our experiments on PMLM (Section~\ref{subsec:pmlm_res}) and SFT (Section~\ref{subsec:sft_result}), we use the pre-trained English RoBERTa\textsubscript{Base}~\cite{liu2019roberta} model as the initial checkpoint model. We use this model, rather than a larger language model, since we run a large number of experiments and needed to be efficient with GPUs. We use the RoBERTa~\footnote{For short, we refer to the official released English RoBERTa\textsubscript{Base} as RoBERTa in the rest of the paper.} tokenizer to process each input sequence and pad or truncate the sequence to a maximal length of $64$ BPE tokens. We continue training RoBERTa with our proposed methods for five epochs with a batch size of $8,192$ and then fine-tune the further trained models on downstream datasets. We provide details about our hyper-parameters in Appendix\ref{subsec:models:hyperparameter}. Our \textbf{baseline (1)} fine-tunes original pre-trained RoBERTa on downstream datsets without any further training. 
Our \textbf{baseline (2)} fine-tunes a SOTA Transformer-based PLM for English tweets, i.e., BERTweet~\cite{nguyen-etal-2020-bertweet}, on downstream datasets. For PMLM experiments, we provide \textbf{baseline (3)}, which further pre-trains RoBERTa on \texttt{Naive-Remove} dataset with the random masking strategy and MLM objectives. We refer to this model as RM-NR. We now present our results.


\vspace{-5pt}
\section{Results and Analysis}
\vspace{-5pt}
 We report performance of our models trained with our PM strategy in Section~\ref{subsec:pmlm_res}, where we investigate two types of pragmatic signals (i.e., hashtag and emoji) and the effect of their locations (anywhere vs. at the end). Section~\ref{subsec:sft_result} shows the results of our SFT method with hashtags and emojis. Moreover, we combine our two proposed methods and compare our models to the SOTA models in Sections~\ref{subsec:combine} and~\ref{subsec:comparisons}, respectively. 
\subsection{PMLM Experiments}\label{subsec:pmlm_res}
\begin{table*}[ht]
\tiny
\centering
\begin{tabular}{@{}lcc:cc:cc:cc:cc:cc|c@{}}
\toprule
\multicolumn{1}{c}{\textbf{Task}}                & \multicolumn{1}{c}{\textbf{RB}} & \multicolumn{1}{c}{\textbf{RM-NR}} & \multicolumn{1}{c}{\textbf{RM-N}} & \multicolumn{1}{c}{\textbf{PM-N}} & \multicolumn{1}{c}{\textbf{RM-HA}} & \multicolumn{1}{c}{\textbf{PM-HA}} & \multicolumn{1}{c}{\textbf{RM-HE}} & \multicolumn{1}{c}{\textbf{PM-HE}} & \multicolumn{1}{c}{\textbf{RM-EA}} & \multicolumn{1}{c}{\textbf{PM-EA}} & \multicolumn{1}{c}{\textbf{RM-EE}} & \multicolumn{1}{c|}{\textbf{PM-EE}}  & \multicolumn{1}{c}{\textbf{BTw}} \\ \midrule
Crisis\textsubscript{Oltea}  & 95.95  &    95.78                           & 95.78                                   & +0.14                                & 95.75                                   & +0.10                                & 95.85                                   & +0.02                                & 95.91                                   & \colorbox{green!10}{\textbf{+0.07}}                                & 95.95                                   & -0.18       & 95.88                         \\

Emo\textsubscript{Moham}     & 77.99    &      \colorbox{green!10}{79.15}                      & \colorbox{green!10}{79.43}                                   & \colorbox{green!10}{+1.30}                                & \colorbox{green!10}{80.31}                                   & \colorbox{green!10}{-1.75}                                & \colorbox{green!10}{79.51}                                   & \colorbox{green!10}{+0.64}                                & \colorbox{green!10}{80.03}                                   & \colorbox{green!10}{+1.06}                                & \colorbox{green!10}{81.28}                                   & \colorbox{green!10}{\textbf{+0.90}}       & 80.14                         \\

Hate\textsubscript{Waseem}   & 57.34    &           57.22                   & 56.75                                   & -0.41                                & 57.16                                   & \colorbox{green!10}{\textbf{+0.35}}                                & 56.97                                   & +0.16                                & 57.00                                   & +0.01                                & 57.08                                   & -0.39           &   57.47                     \\

Hate\textsubscript{David}    & 77.71         &     77.54                       & 77.47                                   & \colorbox{green!10}{\textbf{+0.81}}                                & 76.87                                   & +0.59                                & 77.55                                   & -0.33                                & \colorbox{green!10}{78.13}                                   & \colorbox{green!10}{+0.13}                                & \colorbox{green!10}{78.16}                                   & \colorbox{green!10}{-0.23}           &   77.15                     \\

Humor\textsubscript{Potash}  & 54.40              &   \colorbox{green!10}{54.80}                  & \colorbox{green!10}{55.45}                                   & \colorbox{green!10}{-0.19}                                & \colorbox{green!10}{55.32}                                   & -2.83                                & 50.06                                   & \colorbox{green!10}{+4.54}                                & \colorbox{green!10}{\textbf{57.14}}                                   & \colorbox{green!10}{-2.04}                                & \colorbox{green!10}{55.25}                                   & \colorbox{green!10}{+0.32}             &    52.77                   \\

Humor\textsubscript{Meaney}  & 92.37                &        \colorbox{green!10}{93.50}        & \colorbox{green!10}{93.24}                                   & \colorbox{green!10}{+0.45}                                & \colorbox{green!10}{93.58}                                   & \colorbox{green!10}{-0.10}                                & \colorbox{green!10}{92.85}                                   & \colorbox{green!10}{\textbf{+1.67}}                                & \colorbox{green!10}{93.55}                                   & \colorbox{green!10}{+0.95}                                & \colorbox{green!10}{93.19}                                   & \colorbox{green!10}{-0.50}                & 94.46                \\

Irony\textsubscript{Hee-A}    & 73.93                    &      \colorbox{green!10}{74.46}      & \colorbox{green!10}{74.52}                                   & \colorbox{green!10}{+0.45}                                & \colorbox{green!10}{74.50}                                    & \colorbox{green!10}{+0.66}                                & \colorbox{green!10}{73.97}                                   & \colorbox{green!10}{+2.27}                                & \colorbox{green!10}{75.34}                                   & \colorbox{green!10}{\textbf{+2.59}}                                & \colorbox{green!10}{74.40}                                   & \colorbox{green!10}{+1.22}               &   77.35                 \\

Irony\textsubscript{Hee-B}    & 52.30                    &           50.70       & \colorbox{green!10}{52.91}                                   & \colorbox{green!10}{+0.88}                                & 51.43                                   & -2.14                                & 50.41                                   & \colorbox{green!10}{+4.35}                                & \colorbox{green!10}{54.94}                                   & \colorbox{green!10}{\textbf{+1.15}}                                & \colorbox{green!10}{54.73}                                   & \colorbox{green!10}{-2.26}                     & 58.67           \\
Offense\textsubscript{Zamp} & 80.13          &    \colorbox{green!10}{80.38}                  & 79.97                                   & \colorbox{green!10}{+0.27}                                & 79.74                                   & -0.40                                & 79.95                                   & -1.08                                & \colorbox{green!10}{80.18}                                   & \colorbox{green!10}{\textbf{+0.96}}                                & \colorbox{green!10}{80.18}                                   & \colorbox{green!10}{+0.47}                    &    78.49            \\
Sarc\textsubscript{Riloff}   & 73.85                &            \colorbox{green!10}{73.90}    & 72.02                                   & \colorbox{green!10}{+3.22}                                & 71.42                                   & \colorbox{green!10}{+3.30}                               & \colorbox{green!10}{74.16}                                  & \colorbox{green!10}{+1.72}                               & \colorbox{green!10}{76.52}                                  & \colorbox{green!10}{+1.41}                               & \colorbox{green!10}{76.30}                                  & \colorbox{green!10}{\textbf{+3.80}}            &    78.81                   \\

Sarc\textsubscript{Ptacek}   & 95.09        &         \colorbox{green!10}{96.15}                    & \colorbox{green!10}{95.81}                                  & \colorbox{green!10}{-0.17}                               & \colorbox{green!10}{95.50}                                  & \colorbox{green!10}{+0.12}                               & \colorbox{green!10}{95.24}                                  & \colorbox{green!10}{+0.57}                               & \colorbox{green!10}{95.81}                                  & \colorbox{green!10}{\textbf{+0.25}}                               & \colorbox{green!10}{95.67}                                  & \colorbox{green!10}{+0.34}               &    96.35                \\

Sarc\textsubscript{Rajad}    & 85.07        &       \colorbox{green!10}{85.63}                  & \colorbox{green!10}{86.18}                                   & \colorbox{green!10}{+0.05}                                & 85.04                                   & \colorbox{green!10}{+0.51}                                & \colorbox{green!10}{85.20}                                    & \colorbox{green!10}{+0.73}                                & \colorbox{green!10}{86.14}                                   & \colorbox{green!10}{+0.51}                                & \colorbox{green!10}{86.02}                                   & \colorbox{green!10}{\textbf{+0.92}}            &   87.58                    \\

Sarc\textsubscript{Bam}      & 79.08             &          \colorbox{green!10}{79.27}         & \colorbox{green!10}{80.03}                                   & \colorbox{green!10}{+0.10}                               & \colorbox{green!10}{80.22}                                  & \colorbox{green!10}{-0.06}                               & \colorbox{green!10}{79.83}                                  & \colorbox{green!10}{+0.48}                               & \colorbox{green!10}{80.73}                                  & \colorbox{green!10}{+0.39}                               & \colorbox{green!10}{81.13}                                  & \colorbox{green!10}{\textbf{+0.60}}            &    82.08                    \\

Senti\textsubscript{Rosen}   & 71.08                &           \colorbox{green!10}{71.55}      & \colorbox{green!10}{72.03}                                  & \colorbox{green!10}{\textbf{+0.62}}                               & \colorbox{green!10}{72.10}                                  & \colorbox{green!10}{-0.11}                               & \colorbox{green!10}{71.84}                                  & \colorbox{green!10}{-0.02}                               & \colorbox{green!10}{72.24}                                  & \colorbox{green!10}{-0.26}                               & \colorbox{green!10}{72.27}                                  & \colorbox{green!10}{-0.71}             & 71.83                    \\
Stance\textsubscript{Moham}  & \textbf{70.41}             &   67.00                   & 67.14                                   & +2.80                                & 69.51                                   & -1.38                                & 69.23                                   & +0.45                                & 70.20                                   & -1.58                                & 70.04                                   & -1.56            &   67.41                    \\ \cdashline{1-14} 

\textbf{Average}                                          & 75.78     &        \colorbox{green!10}{75.80}                   & \colorbox{green!10}{75.92}                                   & \colorbox{green!10}{+0.69}                                & \colorbox{green!10}{75.90}                                   & -0.21                                & 75.51                                  & \colorbox{green!10}{+1.08}                                & \colorbox{green!10}{76.92}                                  & \textbf{\colorbox{green!10}{+0.38}}                               & \colorbox{green!10}{76.78}                                  & \colorbox{green!10}{+0.18}         &    77.10                       \\ \bottomrule
\end{tabular} 
\caption{Pragmatic masking results. \textbf{Baselines:} (1) RB: RoBERTa, (2) BTw: BERTweet, (3) RM-NR. \colorbox{green!10}{Light green} indicates our models outperforming the baseline (1).  \textbf{Bold} font indicates best model across all \textit{our} random and pragmatic masking methods. \textbf{\underline{Masking:}} \textbf{RM:} Random masking, \textbf{PM:} Pragmatic masking. \textbf{\underline{Datasets:}} \textbf{N:} \texttt{Naive}, \textbf{NR:} \texttt{Naive-Remove}, \textbf{HA:} \texttt{Hashtag\_any}, \textbf{HE}: \texttt{Hashtag\_end}, \textbf{EA:} \texttt{Emoji\_any}, \textbf{EE:} \texttt{Emoji\_end}. \vspace{-12pt}}\label{tab:pmlm_res} 
\end{table*}  

\textbf{PM on Naive.}~~ We further pre-train RoBERTa on the \texttt{Naive} dataset with our pragmatic masking strategy (PM) and compare to a model trained on the same dataset with random masking (RM). As Table~\ref{tab:pmlm_res} shows, PM-N outperforms RM-N with an average improvement of $0.69$ macro $F_1$ points across the $15$ tasks. 
 We also observe that PM-N improves over RM-N in $12$ out of the $15$ tasks, thus reflecting the effectiveness of our PM strategy even when working with a dataset such as \texttt{Naive} where it is not guaranteed (although likely) that a tweet has hashtags and/or emojis. Moreover, RM-N outperforms RM-NR on eight tasks with improvement of $0.12$ average $F_1$. This indicates that pragmatic cues (i.e., emoji and hashtags) are essential for learning social media data.


\noindent\textbf{PM of Hashtags.}~~ To study the effect of PM on the controlled setting where we guarantee each sample has at least one hashtag \textit{anywhere}, we further pre-train RoBERTa on the \texttt{Hashtag\_any} dataset with PM (PM-HA in Table~\ref{tab:pmlm_res}) and compare to a model further pre-trained on the same dataset with the RM (RM-HA). As Table~\ref{tab:pmlm_res} shows, PM-HA does not improve over RM-HA. Rather, PM-HA results are marginally lower than those of RM-HA. We suspect that the degradation is due to confusions when a hashtag is used as a word of a sentence. Thus, we investigate the effectiveness of hashtag location.


\noindent\textit{\textbf{Effect of Hashtag Location.}}~~ Previous studies~\cite{ren2016context, abdul-2017-emonet} use hashtags as a proxy to label data with social meaning concepts, indicating that hashtags occuring at the end of posts are reliable cues. Hence, we further pre-train RoBERTa on the \texttt{Hashtag\_end} dataset with PM and RM, respectively. As Table~\ref{tab:pmlm_res} shows, PM exploiting hashtags in the end (PM-HE) outperforms random masking (RM-HE) with an average improvement of $1.08$ $F_1$ across the $15$ tasks. It is noteworthy that PM-HE shows improvements over RM-HE in the majority of tasks ($12$ tasks), and both of them outperform the baselines (1) and (3). Compared to RM-HA and PM-HA, the results demonstrate the utility of end-location hashtags on training a LM.


\noindent \textbf{PM of Emojis.}~~ Again, in order to study the impact of PM of emojis under a controlled condition where we guarantee each sample has at least one emoji, we further pre-train RoBERTa on the \texttt{Emoji\_any} dataset with PM and RM, respectively. As Table~\ref{tab:pmlm_res} shows, both methods result in sizable improvements on most of tasks. PM-EA outperforms the random masking method (RM-EA) (macro $F_1$ =$0.38$ improvement) and also exceeds the baseline (1), (2), and (3) with $1.52$, $0.20$, and $1.50$ average $F_1$, respectively. PM-EA thus obtains the best overall performance (macro $F_1$ = $77.30$) and also achieves the best performance on Crisis\textsubscript{Oltea-14}, two irony detection tasks, Offense\textsubscript{Zamp}, and Sarc\textsubscript{Ptacek} across all settings of our PM. 
This indicates that emojis carry important knowledge for social meaning tasks and demonstrates the effectiveness of our PM mechanism to distill and transfer this knowledge to diverse tasks. 

\noindent \textit{\textbf{Effect of Emoji Location.}}~~ We analyze whether learning is sensitive to emoji location: we further pre-train RoBERTa on \texttt{Emoji\_end} dataset with PM and RM and refer to these two models as PM-EE and RM-EE, respectively. Both models perform better than our baselines (1) and (3), and PM-EE achieves the best performance on four datasets across all settings of our PM. Unlike the case of hashtags, the location of the masked emoji is not sensitive for the learning.  

Overall, results show the effectiveness of our PMLM method in improving the self-supervised LM. All models trained with PM on emoji data obtain better performance than those pre-trained on hashtag data. It suggests that emoji cues are somewhat more helpful than hashtag cues for this type of guided model pre-training in the context of social meaning tasks. This implies emojis are more relevant to many social meaning tasks than hashtags are. In other words, in addition to them being cues for social meaning, hashtags can also stand for general topical categories to which different social meaning concepts can apply (e.g., \textit{\#lunch} can be accompanied by both \textit{happy} and \textit{disgust} emotions).

\vspace{-15pt}
\subsection{SFT Experiments}\label{subsec:sft_result}

We conduct SFT using hashtags and emojis. We continue training the original RoBERTa on the \texttt{Hashtag\_pred} and \texttt{Emoji\_pred} dataset for $35$ epochs and refer to these trained models as \textbf{SFT-H} and \textbf{SFT-E}, respectively. 
To evaluate SFT-H and SFT-E, we further fine-tune the obtained models on $15$ task-specific datasets. 
As Table~\ref{tab:sft_res} shows, SFT-E outperforms the first baseline (i.e., RoBERTa) with $1.16$ $F_1$ scores. Comparing SFT-E and PMLM trained with the same dataset (PM-EE), we observe that the two models perform similarly ($76.94$ for SFT-E vs. $76.96$ for PM-EE). Our proposed SFT-H method is also highly effective. On average, SFT-H achieves $2.19$ and $0.87$ $F_1$ improvement over our baseline (1) and (2), respectively. SFT-H also yields sizeable improvements on datasets with smaller training samples, such Irony\textsubscript{Hee-B} (improvement of $7.84$ $F_1$) and Sarc\textsubscript{Riloff} (improvement of $6.65$ $F_1$). Comparing SFT-H to the PMLM model trained with the same dataset (i.e., PM-HE), we observe that SFT-H also outperforms PM-H with $1.38$ $F_1$. This result indicate that SFT can more effectively utilize the information from tweets with hashtags.

\begin{table}[h]
\centering
\tiny
\begin{tabular}{lc:cc:cc|c}
\toprule
\multicolumn{1}{c}{\textbf{Task}}                     & \textbf{RB}    & \textbf{SFT-E} & \textbf{SFT-H} & \multicolumn{1}{c}{\textbf{PragS1}} & \multicolumn{1}{c|}{\textbf{PragS2}} &  \textbf{BTw}  \\ \midrule
Crisis\textsubscript{Oltea} & 95.95                           & \multicolumn{1}{c}{95.76}       & 95.87                           & \textbf{96.02}    & 95.68               &  95.88               \\
Emo\textsubscript{Moham}    & 77.99                           & 79.69                           & 78.69                           & \textbf{82.04}    & 80.50   & 80.14                            \\
Hate\textsubscript{Waseem}  & 57.34                           & 56.47                           & \textbf{63.97} & 60.92                              & 60.25           &  57.47                   \\
Hate\textsubscript{David}   & \textbf{77.71} & 76.45                           & 77.29                           & 77.00                              & 76.93           &   77.15                  \\
Humor\textsubscript{Potash} & 54.40                           & 54.75                           & \textbf{55.51} & 54.93                              & 53.83        & 52.77                      \\
Humor\textsubscript{Meaney} & 92.37                           & 93.82                           & 93.74                           & 93.68                              & \textbf{94.49}    &  94.46 \\
Irony\textsubscript{Hee-A}  & 73.93                           & 76.63                           & 76.22                           & 72.73                              & \textbf{79.89}    &   77.35 \\
Irony\textsubscript{Hee-B}  & 52.30                           & 57.59                           & 60.14                           & 56.11                              & \textbf{61.67}   &   58.67 \\
Offense\textsubscript{Zamp} & 80.13                           & 80.18                           & 79.82                           & \textbf{81.34}    & 79.50         & 78.49                     \\
Sarc\textsubscript{Riloff}  & 73.85                           & 78.34                           & \textbf{80.50} & 78.74                              & 80.49         &  78.81                     \\
Sarc\textsubscript{Ptacek}  & 95.09                           & 95.88                           & 96.01                           & 96.16                              & \textbf{96.24}    &  96.35  \\
Sarc\textsubscript{Rajad}   & 85.07                           & 86.80                           & 87.56                           & 87.48                              & \textbf{88.92}    &  87.58   \\
Sarc\textsubscript{Bam}     & 79.08                           & 81.48                           & 81.19                           & \textbf{82.53}    & 81.53          &  82.08                    \\
Senti\textsubscript{Rosen}  & 71.08                           & 71.27                           & 71.83                           & \textbf{72.07}    & 71.08         &   71.38                     \\
Stance\textsubscript{Moham} & 70.41                           & 69.06                           & \textbf{71.27} & 69.65                              & 70.77          & 67.41                    \\ \cdashline{1-7}
\textbf{Average}      & 75.78                           & 76.94                           & 77.97                           & 77.43                              & \textbf{78.12}    &   77.10\\ \bottomrule
\end{tabular}
\caption{Surrogate fine-tuning (SFT). \textbf{Baselines:} RB (RoBERTa) and BTw (BERTweet). \textbf{SFT-H:} SFT with hashtags. \textbf{SFT-E:} SFT with emojis. \textbf{PragS1:} PMLM with \texttt{Hashtag\_end} (best hashtag PM condition) followed by SFT-E. \textbf{PragS2:} PMLM with \texttt{Emoji\_any} (best emoji PM condition) followed by SFT-H.  }\label{tab:sft_res}  
\end{table}

\vspace{-5pt}
\subsection{Combining PM and SFT}\label{subsec:combine}
To further improve the PMLM with SFT, we take the best hashtag-based model (i.e., PM-HE in Table~\ref{tab:pmlm_res}) and fine-tune on \texttt{Emoji\_pred} (i.e., SFT-E) for $35$ epochs. We refer to this last setting as PM-HE+SFT-E but use the easier alias \textbf{PragS1} in Table~\ref{tab:sft_res}. We observe that PragS1 outperforms both, reaching an average $F_1$ of $77.43$ vs. $75.78$ for the baseline (1) and $76.94$ for SFT-E. Similarly, we also take the best emoji-based PMLM (i.e., PM-EA in Table~\ref{tab:pmlm_res}) and fine-tune on \texttt{Hashtag\_pred} SFT (i.e., SFT-H) for $35$ epochs. This last setting is referred to as PM-EA+SFT-H, but we again use the easier alias \textbf{PragS2}. Our best result is achieved with a combination of PM with emojis and SFT on hashtags (the PragS2 condition). This last model achieves an average $F_1$ of $78.12$ and is $2.34$ and $1.02$ average points higher than baselines of RoBERTa and BERTweet, respectively.

\vspace{-5pt}
\subsection{Model Comparisons}\label{subsec:comparisons}
The purpose of our work is to produce representations effective across all social meaning tasks, rather than a single given task. However, we still compare our best model (i.e., PragS2) on each dataset to the SOTA of that particular dataset and the published results on a Twitter evaluation benchmark~\cite{barbieri-2020-tweeteval}. \textit{All our reported results are an average of three runs}, and we report using the same respective metric adopted by original authors on each dataset. As Table~\ref{tab:compare} shows, our model achieves the best performance on eight out of $15$ datasets. On average, our models are $0.97$ points higher than the closest baseline, i.e., BERTweet. This shows the superiority of our methods, even when compared to models trained simply with MLM with $\sim3 \times$ more data ($850$M tweets for BERTweet vs. only $276$M for our best method). We also note that some SOTA models adopt task-specific approaches and/or require task-specific resources. For example,~\citet{bamman2015contextualized} utilize Stanford sentiment analyzer to identify the sentiment polarity of each word. In addition, task-specific methods can still be combined with our proposed approaches to improve performance on individual tasks.
\begin{table}[ht]
\centering
\scriptsize
\begin{tabular}{@{}llccrcH@{}}
\toprule
\multicolumn{1}{c}{\textbf{Task}}                & \multicolumn{1}{c}{\textbf{Metric}} & \multicolumn{1}{c}{\textbf{SOTA}} & \textbf{TwE}              & \multicolumn{1}{c}{\textbf{BTw}} & \multicolumn{1}{c}{\textbf{\begin{tabular}[c]{@{}c@{}}Ours \\ (PragS2)\end{tabular}}} & \multicolumn{1}{H}{\textbf{Setting}} \\ \midrule
Crisis\textsubscript{Oltea}  & M-$F_1$                           & 95.60\textsuperscript{$\star$}                             & -                         & \textbf{95.88}                                 & 95.68                                 & X1+SFT-E                          \\
Emo\textsubscript{Moham}     & M-$F_1$                           & \multicolumn{1}{c}{-}             & \multicolumn{1}{r}{78.50} & 80.14                                 & \textbf{80.50}                                 & PM-EE                               \\
Hate\textsubscript{Waseem}   & W-$F_1$                             & 73.62\textsuperscript{$\star\star$}                             & -                         & 88.00                                 & \textbf{88.36}                                 & SFT-H                                \\
Hate\textsubscript{David}    & W-$F_1$                             & 90.00\textsuperscript{$\dagger$}                             & -                         & \textbf{91.27}                                 & 91.01                                 & PM-N                                 \\
Humor\textsubscript{Potash}  & M-$F_1$                           & \multicolumn{1}{c}{-}             & -                         & 52.77                                 & \textbf{53.83}                                 & RM-EA                               \\
Humor\textsubscript{Meaney}  & M-$F_1$                           & \multicolumn{1}{c}{\textbf{98.54}$^=$}             & -                         & 94.46                                 &   94.49                                 & PM-HE                                \\
Irony\textsubscript{Hee-A}    & $F^{(i)}$          & 70.50\textsuperscript{$\dagger\dagger$}                             & \multicolumn{1}{r}{65.40} & 71.49                                 & \textbf{76.47}                                 & X2+SFT-H                          \\
Irony\textsubscript{Hee-B}    & M-$F_1$                           & 50.70\textsuperscript{$\dagger\dagger$}                             & -                         & 58.67                                 & \textbf{61.67}                                 & X2+SFT-H                          \\
Offense\textsubscript{-Zamp} & M-$F_1$                           & \textbf{82.90}\textsuperscript{$\ddagger$}                             & \multicolumn{1}{r}{80.50} & 78.49                                 & {79.50}                                 & X1+SFT-E                          \\
Sarc\textsubscript{Riloff}   & $F^{(s)}$          & 51.00\textsuperscript{$\ddagger\ddagger$}                             & -                         & 66.35                                 & \textbf{68.88}                                & SFT-H                                \\
Sarc\textsubscript{Ptacek}   & M-$F_1$                           & 92.37\textsuperscript{$\mathsection$}                             & -                         & \textbf{96.35}                                 & 96.24                                 & X2+SFT-H                          \\
Sarc\textsubscript{Rajad}    & Acc                                 & 92.94\textsuperscript{$\mathsection\mathsection$}                             & -                         & 95.29                                 & \textbf{95.66}                                 & X2+SFT-H                          \\
Sarc\textsubscript{Bam}      & Acc                                 & \textbf{85.10}\textsuperscript{$\|$}                             & -                         & 82.28                                 & 81.27                                & X1+SFT-E                          \\
Senti\textsubscript{Rosen}   & M-Rec                             & 68.50\textsuperscript{$\diamondsuit$}                             & \multicolumn{1}{r}{72.60} & \textbf{72.90}                                 & 71.76                                 & PM-N                                 \\
Stance\textsubscript{Moham}  &
Avg(a,f)             & 71.00\textsuperscript{$\circledcirc$}                             & \multicolumn{1}{r}{69.30} & 69.79                                 & \textbf{73.45}                                 & SFT-H                                \\\cdashline{1-7} 
\textbf{Average}                                          & \multicolumn{1}{c}{-}               & 77.02                             & \multicolumn{1}{r}{73.26} & 79.61                                 & \multicolumn{1}{c}{\textbf{80.58}}                                & \multicolumn{1}{H}{-}                \\ \bottomrule
\end{tabular} 
\caption{Model comparisons. \textbf{SOTA:} Best performance on each respective dataset. \textbf{TwE:} TweetEval~\cite{barbieri-2020-tweeteval} is a benchmark for tweet classification evaluation. \textbf{BTw:} BERTweet~\cite{nguyen-etal-2020-bertweet}. We compare using the same metrics employed on each dataset. \textbf{Metrics:} \textbf{M-$F_1$:} macro $F_1$,  \textbf{W-$F_1$:} weighted $F_1$, $F_1^{(i)}$: $F_1$ irony class, \textbf{$F_1^{(i)}$:} $F_1$ irony class, $F_1^{(s)}$: $F_1$ sarcasm class, \textbf{M-Rec:} macro recall,  \textbf{Avg(a,f)}: Average $F_1$ of the \textit{against} and \textit{in-favor} classes (three-way dataset). \textsuperscript{$\star$}~\citet{liu2020crisisbert}, \textsuperscript{$\star\star$}~\citet{waseem-2016-hateful}, \textsuperscript{$\dagger$}~\citet{davidson-2017-hateoffensive},
\textsuperscript{$^=$}~\citet{meaney2021hahackathon},
\textsuperscript{$\dagger\dagger$}~\citet{van-hee2018semeval}, \textsuperscript{$\ddagger$}~\citet{zampieri-2019-semeval}, \textsuperscript{$\ddagger\ddagger$}~\citet{riloff2013sarcasm}, \textsuperscript{$\mathsection$}~\citet{ptavcek2014sarcasm},  \textsuperscript{$\mathsection\mathsection$}~\citet{rajadesingan2015sarcasm}, \textsuperscript{$\|$}~\citet{bamman2015contextualized}, \textsuperscript{$\diamondsuit$}~\citet{rosenthal-2017-semeval}, \textsuperscript{$\circledcirc$}~\citet{mohammad-2016-semeval}. 
}\label{tab:compare}
\end{table}

\vspace{-3pt}
\section{Zero- and Few-Shot Learning}\label{sec:zfew}
\begin{figure}[!ht]
\centering
\includegraphics[width=\linewidth]{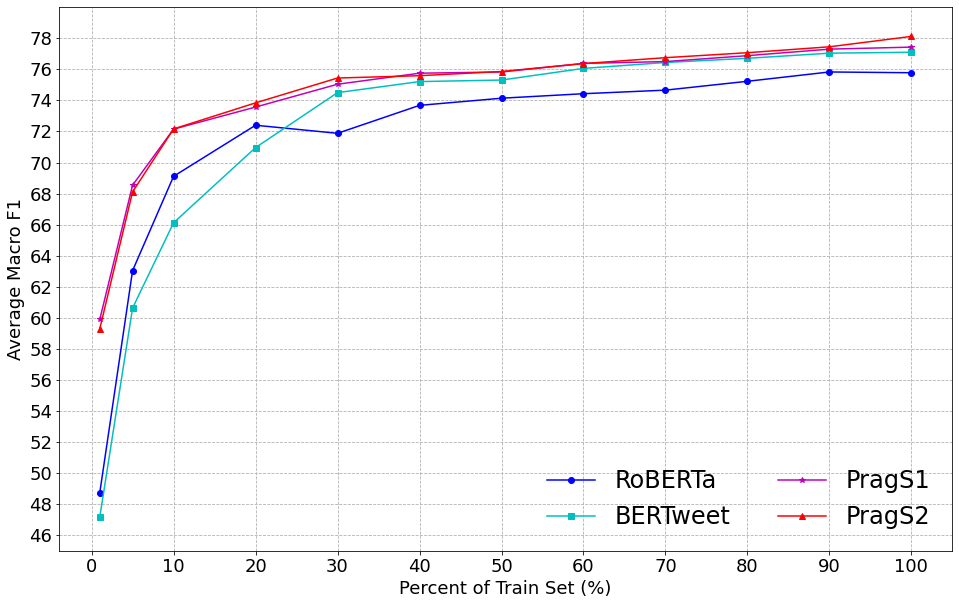} 
\caption{Few-shot learning on downstream with varying percentages of Train sets. The y-axis indicates the average Test macro $F_1$ across the $15$ tasks. The x-axis indicates the percentage of Train set used to fine-tune the model.}\label{fig:data_percent}
\end{figure}

Since our methods exploit general cues in the data for pragmatic masking and learn a broad range of social meaning concepts, we hypothesize they should be particularly effective in \textbf{\textit{few-shot learning}}. To test this hypothesis, we fine-tune our best models (i.e., PragS1 and PragS2) on varying percentages of the Train set of each task as explained in Section~\ref{sec:smpb}. Figure~\ref{fig:data_percent} shows that our two models \textit{always} achieve better average macro $F_1$ scores than each of the RoBERTa and BERTweet baselines across \textit{all} data size settings. Strikingly, our PragS1 and PragS2 outperform RoBERTa with an impressive $11.16$ and $10.55$ average macro $F_1$, respectively, when we fine-tune them on only $1\%$ of the downstream gold data. If we use only $5\%$ of gold data, our PragS1 and  PragS2 improve over the RoBERTa baseline with $5.50\%$ and $5.08$ points, respectively. This demonstrates that our proposed methods most effectively alleviate the challenge of labeled data even under the \textit{severely} few-shot setting. In addition, we observe that the domain-specific LM, BERTweet,  is outperformed by RoBERTa when labeled training data is severely scarce ($\leq 20\%$) (although it achieves superior performance when it is fine-tuned on the full dataset). \textit{These results suggest that, for the scarce data setting, it may be better to further pre-train and surrogate fine-tune an PLM than pre-train a domain-specific LM from scratch.}
We provide model performance on each downstream task and various few-shot settings in Section~\ref{sec:append_fewshot} in Appendix. 

\begin{table}[ht]
\small
\centering
\begin{tabular}{@{}llcc@{}}
\toprule
\multicolumn{2}{c}{\textbf{Task}}      & \textbf{RB}      & \textbf{Prag2} \\ \midrule
\multirow{2}{*}{Arabic}  & Emo\textsubscript{Mageed}     & 29.81          & \textbf{40.37}     \\
                         & Irony\textsubscript{Ghan}       & 31.53          & \textbf{44.40}     \\\cdashline{1-4} 
\multirow{2}{*}{Italian} & Emo\textsubscript{Bian}     & \textbf{27.22} & 26.40              \\
                         & Hate\textsubscript{Bosco} & 40.59          & \textbf{47.04}     \\\cdashline{1-4} 
\multirow{2}{*}{Spanish} & Emo\textsubscript{Moham}     & 30.58          & \textbf{35.09}     \\
                         & Hate\textsubscript{Bas} & 41.43          & \textbf{43.66}     \\ \cdashline{1-4} 
\multicolumn{2}{c}{\textbf{Average}}             & 33.53            & \textbf{39.49}       \\ \bottomrule
\end{tabular}
\caption{Zero-shot performance. \textbf{RB:} RoBERTa.}\label{tab:zero}
\end{table}

Our proposed methods are language agnostic, and may fare well on languages other than English. Although we do not test this claim directly in this work, we do score our English-language best models on six datasets from three other languages (\textit{\textbf{zero-shot setting}}). 
We fine-tune our best English model (i.e., PragS2 in Table~\ref{tab:sft_res}) on the English dataset Emo\textsubscript{Moham}, Irony\textsubscript{Hee-A}, and Hate\textsubscript{David} and, then, evaluate on the Test set of emotion, irony, and hate speech datasets from other languages, respectively. We compare these models against the English RoBERTa baseline fine-tuned on the same English data. As Table~\ref{tab:zero} shows, our models outperform the baseline in the zero-shot setting on five out of six dataset with an average improvement of $5.96$ $F_1$. These results emphasize the effectiveness of our methods even in the zero-shot setting across different languages and tasks, and motivate future work further extending our methods to other languages.

\section{Model Analyses}
 To better understand model behavior, we carry out both a qualitative and a quantitative analysis. For the qualitative analysis, we encode all the Dev and Test samples from one emotion downstream task using two PLMs (RoBERTa and BERTweet) and our two best models (i.e., PragS1 and PragS2)\footnote{Note that we use these representation models \textit{without} downstream fine-tuning.}. We then use the hidden state of the [CLS] token from the last Transformer encoder layer as the representation of each input. We then map these tweet representation vectors ($768$ dimensions) to a $2$-D space through t-SNE technique~\cite{van2008visualizing} and visualize the results. Comparing our models to the original RoBERTa and BERTweet, we observe that the representations from our models give sensible clustering of emotions before fine-tuning on downstream dataset. 
 
 \begin{figure}[h]
\centering
\begin{subfigure}[b]{.235\textwidth}
  \centering
  \includegraphics[width=\textwidth]{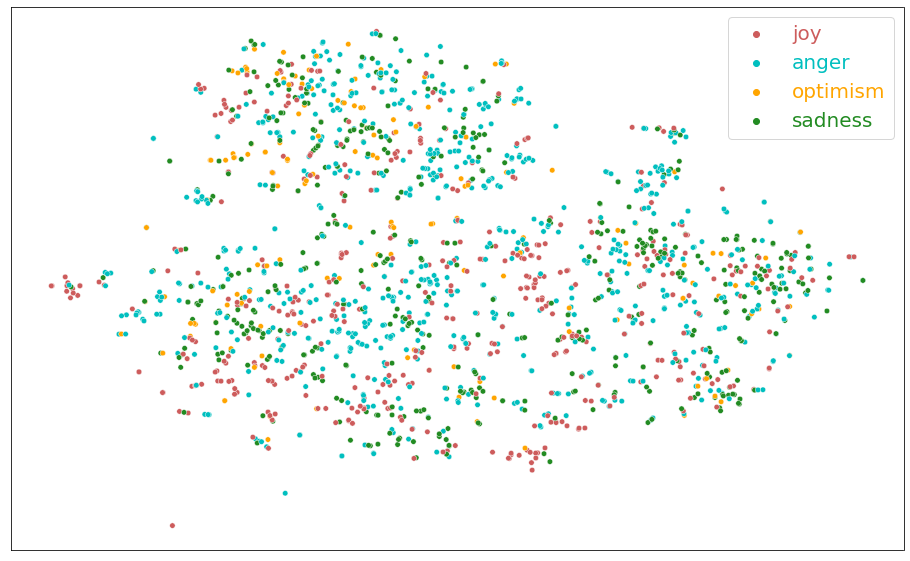}
  \caption{RoBERTa}
\end{subfigure}
\hfill
\begin{subfigure}[b]{.235\textwidth}
  \centering
  \includegraphics[width=\textwidth]{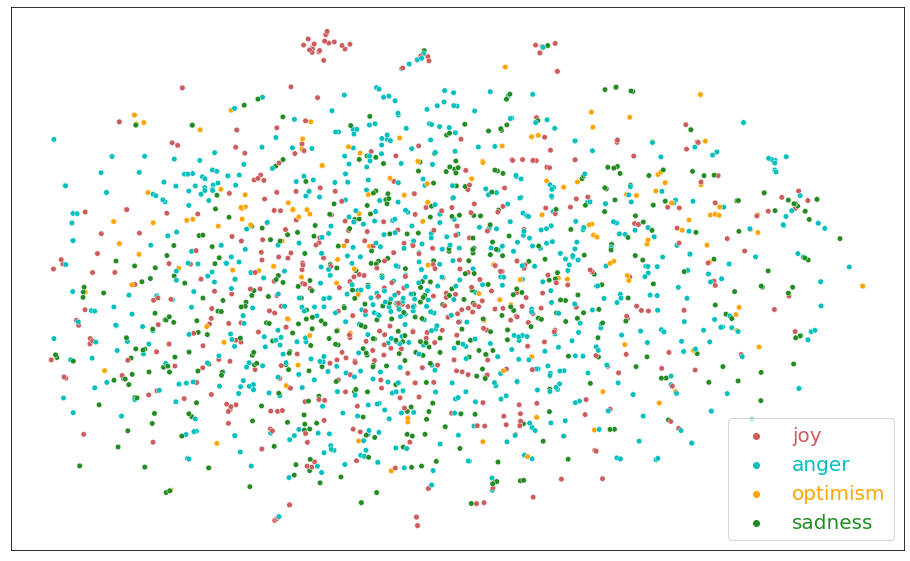}
  \caption{BERTweet}
\end{subfigure}
\hfill
\begin{subfigure}[b]{.235\textwidth}
  \centering
  \includegraphics[width=\textwidth]{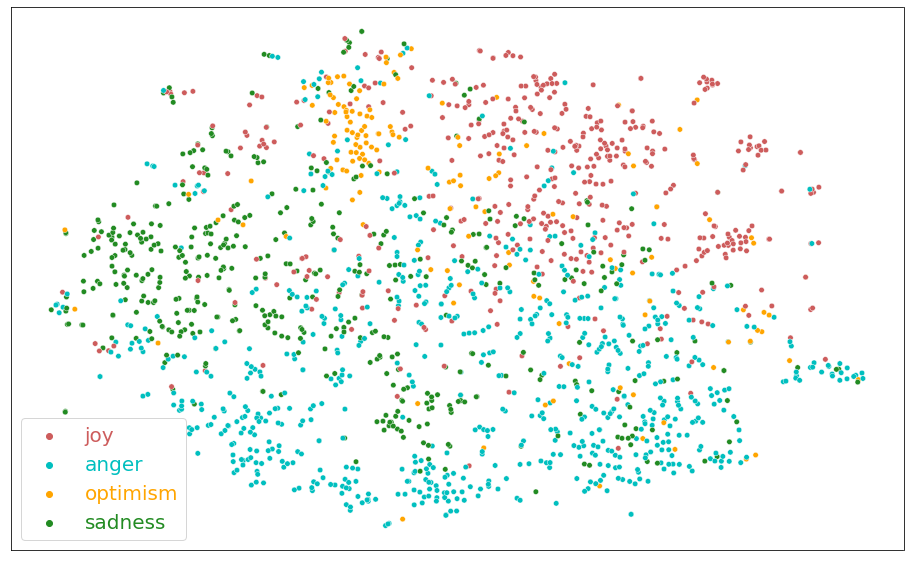}
  \caption{PragS1}
\end{subfigure}
\hfill
\begin{subfigure}[b]{.235\textwidth}
  \centering
  \includegraphics[width=\textwidth]{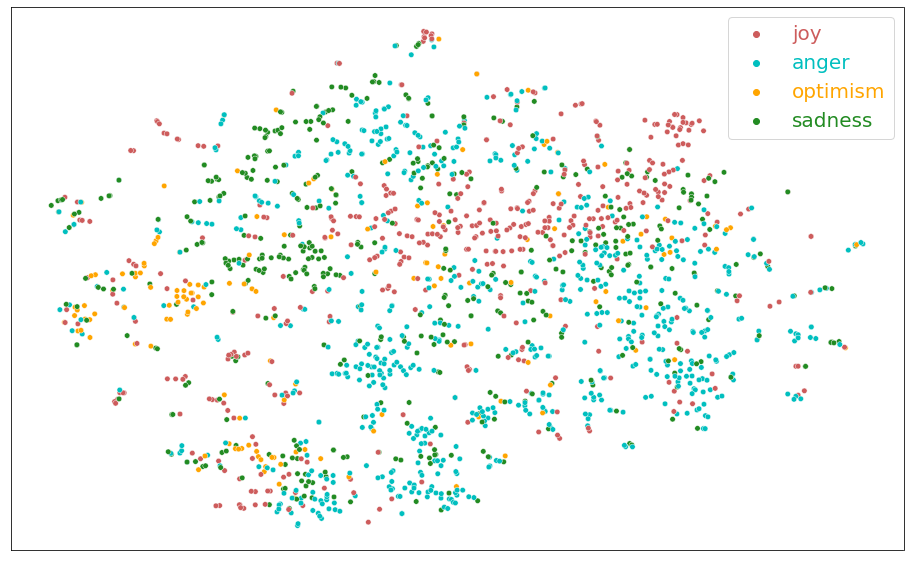}
  \caption{PragS2}
\end{subfigure}
     
\caption{t-SNE plots of the learned embeddings on Dev and Test sets of Emo\textsubscript{Moham}. Our learned representations clearly help tease apart the different classes.}
\label{fig:tsne_task}
\end{figure}

Recent research~\cite{ethayarajh-2019-contextual, li-etal-2020-sentence, gao-etal-2021-simcse} has identified an anisotropy problem with the sentence embedding from PLMs, i.e., learned representations occupy a narrow cone, which significantly undermines their expressiveness. Hence, several concurrent studies~\cite{gao-etal-2021-simcse, liu2021fast} seek to improve uniformity of PLMs. However,~\citet{wang-2021-understanding} reveal a uniformity-tolerance dilemma, where excessive uniformity makes a model intolerant to semantically similar samples, thereby breaking its underlying semantic structure. Following~\citet{wang-2021-understanding}, we investigate the uniformity and tolerance of our models. The uniformity metric indicates the embedding distribution in a unit hypersphere, and the tolerance metric is the mean similarities of samples belonging to the same class. Formulas of uniformity and tolerance are defined in Section~\ref{sec:uni-tole} in appendix. We calculate these two metrics for each model using development data from our $13$ downstream datasets (excluding Crisis\textsubscript{Oltea} and Stance\textsubscript{Moham}). As Table~\ref{tab:uniformity-tolerance} shows, RoBERTa obtains a low uniformity and high tolerance score with its representations are located at a narrow cone where the cosine similarities of data points are extremely high. Results reveal that none of MLMs (i.e., pragmatic masking and random masking models) improves the spatial anisotropy. Nevertheless, surrogate fine-tuning is able to alleviate the anisotropy improving the uniformity. SFT-H achieves best uniformity (at $3.00$). Our hypothesis is that fine-tuning on our extremely fine-grained hashtag prediction task forces the model to learn a more uniform representation where hashtag classes are separable. Finally, we observe that our best model, Prag2, makes a balance between uniformity and tolerance (uniformity$=2.36$, tolerance$=0.35$). 
 \begin{table}[h]
\centering
\scriptsize
\begin{tabular}{@{}lccc@{}}
\toprule
\multicolumn{1}{c}{\textbf{Model}} & \textbf{Performance} & \textbf{Uniformity} & \textbf{Tolerance} \\ \midrule
RoBERTa                            & 75.78                & 0.02                & \textbf{1.00}               \\ \cdashline{1-4}
RM-NR                              & 75.80                & 0.06                & 0.99               \\
RM-N                               & 75.92                & 0.06                & 0.99               \\
PM-N                               & 76.61                & 0.02                & 0.99               \\
RM-HA                              & 75.90                & 0.01                & 0.99               \\
PM-HA                              & 75.69                & 0.04                & 0.99               \\
RM-HE                              & 75.51                & 0.02                & 0.99               \\
PM-HE                              & 76.59                & 0.05                & 0.99               \\
RM-EA                              & 76.92                & 0.02                & 1.00               \\
PM-EA                              & 77.30                & 0.02                & 0.99               \\
RM-EE                              & 76.78                & 0.02                & 0.99               \\
PM-EE                              & 76.96                & 0.03                & 0.99               \\ \hline
SFT-H                              & 77.79                & \textbf{3.00}                & 0.21               \\
SFT-E                              & 76.94                & 2.65                & 0.30               \\ \cdashline{1-4}
PragS1                             & 77.43                & 2.98                & 0.21               \\
PragS2                             & \textbf{78.12}                & 2.36                & 0.35               \\ \bottomrule
\end{tabular}
\caption{Comparison of uniformity and tolerance. For both metrics, higher is better. }\label{tab:uniformity-tolerance}
\end{table}

\section{Conclusion}\label{sec:conclude} 
We proposed two novel methods for improving transfer learning with PLMs, pragmatic masking and surrogate fine-tuning, and demonstrated the effectiveness of these methods on a wide range of social meaning datasets. 
Our models exhibit remarkable performance in the few-shot setting and even the severely few-shot setting. Our models also establish new SOTA on eight out of fifteen datasets when compared to tailored, task-specific models with access to external resources. Our proposed methods are also language independent, and show promising performance when applied in zero-shot settings on six datasets from three different languages. In future research, we plan to further test this language independence claim. We hope our methods will inspire new work on improving language models without use of much labeled data.

\section*{Acknowledgements}\label{sec:acknow}
We gratefully acknowledge support from the Natural Sciences and Engineering Research Council of Canada (NSERC; RGPIN-2018-04267), the Social Sciences and Humanities Research Council of Canada (SSHRC; 435-2018-0576; 895-2021-1008), Canadian Foundation for Innovation (CFI; 37771), Compute Canada (CC),\footnote{\href{https://www.computecanada.ca}{https://www.computecanada.ca}}, and UBC ARC-Sockeye.\footnote{\href{https://arc.ubc.ca/ubc-arc-sockeye}{https://arc.ubc.ca/ubc-arc-sockeye}} Any opinions, conclusions or recommendations expressed in this material are those of the author(s) and do not necessarily reflect the views of NSERC, SSHRC, CFI, CC, or UBC ARC-Sockeye. We thank AbdelRahim ElMadany for help with data preparation. 
\bibliography{custom}

\begin{thebibliography}{58}
\expandafter\ifx\csname natexlab\endcsname\relax\def\natexlab#1{#1}\fi

\bibitem[{Abdul-Mageed and Ungar(2017)}]{abdul-2017-emonet}
Muhammad Abdul-Mageed and Lyle Ungar. 2017.
\newblock \href {https://doi.org/10.18653/v1/P17-1067} {{E}mo{N}et:
  Fine-grained emotion detection with gated recurrent neural networks}.
\newblock In \emph{Proceedings of the 55th Annual Meeting of the Association
  for Computational Linguistics (Volume 1: Long Papers)}, pages 718--728,
  Vancouver, Canada. Association for Computational Linguistics.

\bibitem[{Abdul-Mageed et~al.(2020)Abdul-Mageed, Zhang, Hashemi, and
  Nagoudi}]{mageed-2020-aranet}
Muhammad Abdul-Mageed, Chiyu Zhang, Azadeh Hashemi, and El~Moatez~Billah
  Nagoudi. 2020.
\newblock \href {https://aclanthology.org/2020.osact-1.3} {{A}ra{N}et: A deep
  learning toolkit for {A}rabic social media}.
\newblock In \emph{Proceedings of the 4th Workshop on Open-Source Arabic
  Corpora and Processing Tools, with a Shared Task on Offensive Language
  Detection}, pages 16--23, Marseille, France. European Language Resource
  Association.

\bibitem[{Bamman and Smith(2015)}]{bamman2015contextualized}
David Bamman and Noah~A. Smith. 2015.
\newblock \href
  {http://www.aaai.org/ocs/index.php/ICWSM/ICWSM15/paper/view/10538}
  {Contextualized sarcasm detection on twitter}.
\newblock In \emph{Proceedings of the Ninth International Conference on Web and
  Social Media, {ICWSM} 2015, University of Oxford, Oxford, UK, May 26-29,
  2015}, pages 574--577. {AAAI} Press.

\bibitem[{Barbieri et~al.(2020)Barbieri, Camacho-Collados, Espinosa~Anke, and
  Neves}]{barbieri-2020-tweeteval}
Francesco Barbieri, Jose Camacho-Collados, Luis Espinosa~Anke, and Leonardo
  Neves. 2020.
\newblock \href {https://doi.org/10.18653/v1/2020.findings-emnlp.148}
  {{T}weet{E}val: Unified benchmark and comparative evaluation for tweet
  classification}.
\newblock In \emph{Findings of the Association for Computational Linguistics:
  EMNLP 2020}, pages 1644--1650, Online. Association for Computational
  Linguistics.

\bibitem[{Barbieri et~al.(2018)Barbieri, Camacho-Collados, Ronzano,
  Espinosa-Anke, Ballesteros, Basile, Patti, and
  Saggion}]{barbieri-2018-semeval}
Francesco Barbieri, Jose Camacho-Collados, Francesco Ronzano, Luis
  Espinosa-Anke, Miguel Ballesteros, Valerio Basile, Viviana Patti, and Horacio
  Saggion. 2018.
\newblock \href {https://doi.org/10.18653/v1/S18-1003} {{S}em{E}val 2018 task
  2: Multilingual emoji prediction}.
\newblock In \emph{Proceedings of The 12th International Workshop on Semantic
  Evaluation}, pages 24--33, New Orleans, Louisiana. Association for
  Computational Linguistics.

\bibitem[{Basile et~al.(2019)Basile, Bosco, Fersini, Nozza, Patti,
  Rangel~Pardo, Rosso, and Sanguinetti}]{basile-2019-semeval}
Valerio Basile, Cristina Bosco, Elisabetta Fersini, Debora Nozza, Viviana
  Patti, Francisco~Manuel Rangel~Pardo, Paolo Rosso, and Manuela Sanguinetti.
  2019.
\newblock \href {https://doi.org/10.18653/v1/S19-2007} {{S}em{E}val-2019 task
  5: Multilingual detection of hate speech against immigrants and women in
  {T}witter}.
\newblock In \emph{Proceedings of the 13th International Workshop on Semantic
  Evaluation}, pages 54--63, Minneapolis, Minnesota, USA. Association for
  Computational Linguistics.

\bibitem[{Bianchi et~al.(2021)Bianchi, Nozza, and Hovy}]{bianchi2021feel}
Federico Bianchi, Debora Nozza, and Dirk Hovy. 2021.
\newblock \href {https://aclanthology.org/2021.wassa-1.8/} {{FEEL-IT:} emotion
  and sentiment classification for the italian language}.
\newblock In \emph{Proceedings of the Eleventh Workshop on Computational
  Approaches to Subjectivity, Sentiment and Social Media Analysis, WASSA@EACL
  2021, Online, April 19, 2021}, pages 76--83. Association for Computational
  Linguistics.

\bibitem[{Bosco et~al.(2018)Bosco, Dell'Orletta, Poletto, Sanguinetti, and
  Tesconi}]{bosco2018overview}
Cristina Bosco, Felice Dell'Orletta, Fabio Poletto, Manuela Sanguinetti, and
  Maurizio Tesconi. 2018.
\newblock \href {http://ceur-ws.org/Vol-2263/paper010.pdf} {Overview of the
  {EVALITA} 2018 hate speech detection task}.
\newblock In \emph{Proceedings of the Sixth Evaluation Campaign of Natural
  Language Processing and Speech Tools for Italian. Final Workshop {(EVALITA}
  2018) co-located with the Fifth Italian Conference on Computational
  Linguistics (CLiC-it 2018), Turin, Italy, December 12-13, 2018}, volume 2263
  of \emph{{CEUR} Workshop Proceedings}. CEUR-WS.org.

\bibitem[{Chang and Lu(2021)}]{chang-2021-rethinking}
Ting-Yun Chang and Chi-Jen Lu. 2021.
\newblock \href {https://doi.org/10.18653/v1/2021.findings-emnlp.61}
  {Rethinking why intermediate-task fine-tuning works}.
\newblock In \emph{Findings of the Association for Computational Linguistics:
  EMNLP 2021}, pages 706--713, Punta Cana, Dominican Republic. Association for
  Computational Linguistics.

\bibitem[{Corazza et~al.(2020)Corazza, Menini, Cabrio, Tonelli, and
  Villata}]{corazza-2020-hybrid}
Michele Corazza, Stefano Menini, Elena Cabrio, Sara Tonelli, and Serena
  Villata. 2020.
\newblock \href {https://doi.org/10.18653/v1/2020.findings-emnlp.84} {Hybrid
  emoji-based masked language models for zero-shot abusive language detection}.
\newblock In \emph{Findings of the Association for Computational Linguistics:
  EMNLP 2020}, pages 943--949, Online. Association for Computational
  Linguistics.

\bibitem[{Davidson et~al.(2017)Davidson, Warmsley, Macy, and
  Weber}]{davidson-2017-hateoffensive}
Thomas Davidson, Dana Warmsley, Michael~W. Macy, and Ingmar Weber. 2017.
\newblock \href {https://aaai.org/ocs/index.php/ICWSM/ICWSM17/paper/view/15665}
  {Automated hate speech detection and the problem of offensive language}.
\newblock In \emph{Proceedings of the Eleventh International Conference on Web
  and Social Media, {ICWSM} 2017, Montr{\'{e}}al, Qu{\'{e}}bec, Canada, May
  15-18, 2017}, pages 512--515. {AAAI} Press.

\bibitem[{Devlin et~al.(2019)Devlin, Chang, Lee, and
  Toutanova}]{devlin-2019-bert}
Jacob Devlin, Ming-Wei Chang, Kenton Lee, and Kristina Toutanova. 2019.
\newblock \href {https://doi.org/10.18653/v1/N19-1423} {{BERT}: Pre-training of
  deep bidirectional transformers for language understanding}.
\newblock In \emph{Proceedings of the 2019 Conference of the North {A}merican
  Chapter of the Association for Computational Linguistics: Human Language
  Technologies, Volume 1 (Long and Short Papers)}, pages 4171--4186,
  Minneapolis, Minnesota. Association for Computational Linguistics.

\bibitem[{Ethayarajh(2019)}]{ethayarajh-2019-contextual}
Kawin Ethayarajh. 2019.
\newblock \href {https://doi.org/10.18653/v1/D19-1006} {How contextual are
  contextualized word representations? {C}omparing the geometry of {BERT},
  {ELM}o, and {GPT}-2 embeddings}.
\newblock In \emph{Proceedings of the 2019 Conference on Empirical Methods in
  Natural Language Processing and the 9th International Joint Conference on
  Natural Language Processing (EMNLP-IJCNLP)}, pages 55--65, Hong Kong, China.
  Association for Computational Linguistics.

\bibitem[{Felbo et~al.(2017)Felbo, Mislove, S{\o}gaard, Rahwan, and
  Lehmann}]{felbo-2017-using}
Bjarke Felbo, Alan Mislove, Anders S{\o}gaard, Iyad Rahwan, and Sune Lehmann.
  2017.
\newblock \href {https://doi.org/10.18653/v1/D17-1169} {Using millions of emoji
  occurrences to learn any-domain representations for detecting sentiment,
  emotion and sarcasm}.
\newblock In \emph{Proceedings of the 2017 Conference on Empirical Methods in
  Natural Language Processing}, pages 1615--1625, Copenhagen, Denmark.
  Association for Computational Linguistics.

\bibitem[{Gao et~al.(2021)Gao, Yao, and Chen}]{gao-etal-2021-simcse}
Tianyu Gao, Xingcheng Yao, and Danqi Chen. 2021.
\newblock \href {https://doi.org/10.18653/v1/2021.emnlp-main.552} {{S}im{CSE}:
  Simple contrastive learning of sentence embeddings}.
\newblock In \emph{Proceedings of the 2021 Conference on Empirical Methods in
  Natural Language Processing}, pages 6894--6910, Online and Punta Cana,
  Dominican Republic. Association for Computational Linguistics.

\bibitem[{Ghanem et~al.(2019)Ghanem, Karoui, Benamara, Moriceau, and
  Rosso}]{idat2019}
Bilal Ghanem, Jihen Karoui, Farah Benamara, V{\'{e}}ronique Moriceau, and Paolo
  Rosso. 2019.
\newblock \href {https://doi.org/10.1145/3368567.3368585} {{IDAT} at
  {FIRE2019:} overview of the track on irony detection in arabic tweets}.
\newblock In \emph{{FIRE} '19: Forum for Information Retrieval Evaluation,
  Kolkata, India, December, 2019}, pages 10--13. {ACM}.

\bibitem[{Gu et~al.(2020)Gu, Zhang, Wang, Liu, and Sun}]{gu-2020-train}
Yuxian Gu, Zhengyan Zhang, Xiaozhi Wang, Zhiyuan Liu, and Maosong Sun. 2020.
\newblock \href {https://doi.org/10.18653/v1/2020.emnlp-main.566} {Train no
  evil: Selective masking for task-guided pre-training}.
\newblock In \emph{Proceedings of the 2020 Conference on Empirical Methods in
  Natural Language Processing (EMNLP)}, pages 6966--6974, Online. Association
  for Computational Linguistics.

\bibitem[{Kawintiranon and Singh(2021)}]{kawintiranon-2021-knowledge}
Kornraphop Kawintiranon and Lisa Singh. 2021.
\newblock \href {https://doi.org/10.18653/v1/2021.naacl-main.376} {Knowledge
  enhanced masked language model for stance detection}.
\newblock In \emph{Proceedings of the 2021 Conference of the North American
  Chapter of the Association for Computational Linguistics: Human Language
  Technologies}, pages 4725--4735, Online. Association for Computational
  Linguistics.

\bibitem[{Ke et~al.(2020)Ke, Ji, Liu, Zhu, and Huang}]{ke-2020-sentilare}
Pei Ke, Haozhe Ji, Siyang Liu, Xiaoyan Zhu, and Minlie Huang. 2020.
\newblock \href {https://doi.org/10.18653/v1/2020.emnlp-main.567}
  {{S}enti{LARE}: Sentiment-aware language representation learning with
  linguistic knowledge}.
\newblock In \emph{Proceedings of the 2020 Conference on Empirical Methods in
  Natural Language Processing (EMNLP)}, pages 6975--6988, Online. Association
  for Computational Linguistics.

\bibitem[{Li et~al.(2020)Li, Zhou, He, Wang, Yang, and
  Li}]{li-etal-2020-sentence}
Bohan Li, Hao Zhou, Junxian He, Mingxuan Wang, Yiming Yang, and Lei Li. 2020.
\newblock \href {https://doi.org/10.18653/v1/2020.emnlp-main.733} {On the
  sentence embeddings from pre-trained language models}.
\newblock In \emph{Proceedings of the 2020 Conference on Empirical Methods in
  Natural Language Processing (EMNLP)}, pages 9119--9130, Online. Association
  for Computational Linguistics.

\bibitem[{Lin et~al.(2021)Lin, Miller, Dligach, Bethard, and
  Savova}]{lin-2021-entitybert}
Chen Lin, Timothy Miller, Dmitriy Dligach, Steven Bethard, and Guergana Savova.
  2021.
\newblock \href {https://doi.org/10.18653/v1/2021.bionlp-1.21} {{E}ntity{BERT}:
  Entity-centric masking strategy for model pretraining for the clinical
  domain}.
\newblock In \emph{Proceedings of the 20th Workshop on Biomedical Language
  Processing}, pages 191--201, Online. Association for Computational
  Linguistics.

\bibitem[{Liu et~al.(2021{\natexlab{a}})Liu, Vulic, Korhonen, and
  Collier}]{liu2021fast}
Fangyu Liu, Ivan Vulic, Anna Korhonen, and Nigel Collier. 2021{\natexlab{a}}.
\newblock \href {https://aclanthology.org/2021.emnlp-main.109} {Fast,
  effective, and self-supervised: Transforming masked language models into
  universal lexical and sentence encoders}.
\newblock In \emph{Proceedings of the 2021 Conference on Empirical Methods in
  Natural Language Processing, {EMNLP} 2021, Virtual Event / Punta Cana,
  Dominican Republic, 7-11 November, 2021}, pages 1442--1459. Association for
  Computational Linguistics.

\bibitem[{Liu et~al.(2021{\natexlab{b}})Liu, Singhal, Blessing, Wood, and
  Lim}]{liu2020crisisbert}
Junhua Liu, Trisha Singhal, Luci{\"{e}}nne T.~M. Blessing, Kristin~L. Wood, and
  Kwan~Hui Lim. 2021{\natexlab{b}}.
\newblock \href {https://doi.org/10.1145/3465336.3475117} {Crisisbert: {A}
  robust transformer for crisis classification and contextual crisis
  embedding}.
\newblock In \emph{{HT} '21: 32nd {ACM} Conference on Hypertext and Social
  Media, Virtual Event, Ireland, 30 August 2021 - 2 September 2021}, pages
  133--141. {ACM}.

\bibitem[{Liu et~al.(2019)Liu, Ott, Goyal, Du, Joshi, Chen, Levy, Lewis,
  Zettlemoyer, and Stoyanov}]{liu2019roberta}
Yinhan Liu, Myle Ott, Naman Goyal, Jingfei Du, Mandar Joshi, Danqi Chen, Omer
  Levy, Mike Lewis, Luke Zettlemoyer, and Veselin Stoyanov. 2019.
\newblock \href {http://arxiv.org/abs/1907.11692} {Roberta: {A} robustly
  optimized {BERT} pretraining approach}.
\newblock \emph{CoRR}, abs/1907.11692.

\bibitem[{Loshchilov and Hutter(2019)}]{loshchilov2018decoupled}
Ilya Loshchilov and Frank Hutter. 2019.
\newblock \href {https://openreview.net/forum?id=Bkg6RiCqY7} {Decoupled weight
  decay regularization}.
\newblock In \emph{7th International Conference on Learning Representations,
  {ICLR} 2019, New Orleans, LA, USA, May 6-9, 2019}. OpenReview.net.

\bibitem[{Meaney et~al.(2021)Meaney, Wilson, Chiruzzo, Lopez, and
  Magdy}]{meaney2021hahackathon}
J.~A. Meaney, Steven~R. Wilson, Luis Chiruzzo, Adam Lopez, and Walid Magdy.
  2021.
\newblock \href {https://doi.org/10.18653/v1/2021.semeval-1.9} {Semeval 2021
  task 7: Hahackathon, detecting and rating humor and offense}.
\newblock In \emph{Proceedings of the 15th International Workshop on Semantic
  Evaluation, SemEval@ACL/IJCNLP 2021, Virtual Event / Bangkok, Thailand,
  August 5-6, 2021}, pages 105--119. Association for Computational Linguistics.

\bibitem[{Mohammad et~al.(2018)Mohammad, Bravo-Marquez, Salameh, and
  Kiritchenko}]{mohammad-2018-semeval}
Saif Mohammad, Felipe Bravo-Marquez, Mohammad Salameh, and Svetlana
  Kiritchenko. 2018.
\newblock \href {https://doi.org/10.18653/v1/S18-1001} {{S}em{E}val-2018 task
  1: Affect in tweets}.
\newblock In \emph{Proceedings of The 12th International Workshop on Semantic
  Evaluation}, pages 1--17, New Orleans, Louisiana. Association for
  Computational Linguistics.

\bibitem[{Mohammad et~al.(2016)Mohammad, Kiritchenko, Sobhani, Zhu, and
  Cherry}]{mohammad-2016-semeval}
Saif Mohammad, Svetlana Kiritchenko, Parinaz Sobhani, Xiaodan Zhu, and Colin
  Cherry. 2016.
\newblock \href {https://doi.org/10.18653/v1/S16-1003} {{S}em{E}val-2016 task
  6: Detecting stance in tweets}.
\newblock In \emph{Proceedings of the 10th International Workshop on Semantic
  Evaluation ({S}em{E}val-2016)}, pages 31--41, San Diego, California.
  Association for Computational Linguistics.

\bibitem[{Nguyen et~al.(2020)Nguyen, Vu, and
  Tuan~Nguyen}]{nguyen-etal-2020-bertweet}
Dat~Quoc Nguyen, Thanh Vu, and Anh Tuan~Nguyen. 2020.
\newblock \href {https://doi.org/10.18653/v1/2020.emnlp-demos.2} {{BERT}weet: A
  pre-trained language model for {E}nglish tweets}.
\newblock In \emph{Proceedings of the 2020 Conference on Empirical Methods in
  Natural Language Processing: System Demonstrations}, pages 9--14, Online.
  Association for Computational Linguistics.

\bibitem[{Olteanu et~al.(2014)Olteanu, Castillo, Diaz, and
  Vieweg}]{olteanu2014crisislex}
Alexandra Olteanu, Carlos Castillo, Fernando Diaz, and Sarah Vieweg. 2014.
\newblock \href
  {http://www.aaai.org/ocs/index.php/ICWSM/ICWSM14/paper/view/8091} {Crisislex:
  {A} lexicon for collecting and filtering microblogged communications in
  crises}.
\newblock In \emph{Proceedings of the Eighth International Conference on
  Weblogs and Social Media, {ICWSM} 2014, Ann Arbor, Michigan, USA, June 1-4,
  2014}. The {AAAI} Press.

\bibitem[{Phang et~al.(2020)Phang, Calixto, Htut, Pruksachatkun, Liu, Vania,
  Kann, and Bowman}]{phang-2020-english}
Jason Phang, Iacer Calixto, Phu~Mon Htut, Yada Pruksachatkun, Haokun Liu, Clara
  Vania, Katharina Kann, and Samuel~R. Bowman. 2020.
\newblock \href {https://aclanthology.org/2020.aacl-main.56} {{E}nglish
  intermediate-task training improves zero-shot cross-lingual transfer too}.
\newblock In \emph{Proceedings of the 1st Conference of the Asia-Pacific
  Chapter of the Association for Computational Linguistics and the 10th
  International Joint Conference on Natural Language Processing}, pages
  557--575, Suzhou, China. Association for Computational Linguistics.

\bibitem[{Potash et~al.(2017)Potash, Romanov, and
  Rumshisky}]{potash-2017-semeval}
Peter Potash, Alexey Romanov, and Anna Rumshisky. 2017.
\newblock \href {https://doi.org/10.18653/v1/S17-2004} {{S}em{E}val-2017 task
  6: {\#}{H}ashtag{W}ars: Learning a sense of humor}.
\newblock In \emph{Proceedings of the 11th International Workshop on Semantic
  Evaluation ({S}em{E}val-2017)}, pages 49--57, Vancouver, Canada. Association
  for Computational Linguistics.

\bibitem[{Poth et~al.(2021)Poth, Pfeiffer, R{\"u}ckl{\'e}, and
  Gurevych}]{poth-2021-pre}
Clifton Poth, Jonas Pfeiffer, Andreas R{\"u}ckl{\'e}, and Iryna Gurevych. 2021.
\newblock \href {https://doi.org/10.18653/v1/2021.emnlp-main.827} {{W}hat to
  pre-train on? {E}fficient intermediate task selection}.
\newblock In \emph{Proceedings of the 2021 Conference on Empirical Methods in
  Natural Language Processing}, pages 10585--10605, Online and Punta Cana,
  Dominican Republic. Association for Computational Linguistics.

\bibitem[{Pruksachatkun et~al.(2020)Pruksachatkun, Phang, Liu, Htut, Zhang,
  Pang, Vania, Kann, and Bowman}]{pruksachatkun-2020-intermediate}
Yada Pruksachatkun, Jason Phang, Haokun Liu, Phu~Mon Htut, Xiaoyi Zhang,
  Richard~Yuanzhe Pang, Clara Vania, Katharina Kann, and Samuel~R. Bowman.
  2020.
\newblock \href {https://doi.org/10.18653/v1/2020.acl-main.467}
  {Intermediate-task transfer learning with pretrained language models: When
  and why does it work?}
\newblock In \emph{Proceedings of the 58th Annual Meeting of the Association
  for Computational Linguistics}, pages 5231--5247, Online. Association for
  Computational Linguistics.

\bibitem[{Pt{\'a}{\v{c}}ek et~al.(2014)Pt{\'a}{\v{c}}ek, Habernal, and
  Hong}]{ptavcek2014sarcasm}
Tom{\'a}{\v{s}} Pt{\'a}{\v{c}}ek, Ivan Habernal, and Jun Hong. 2014.
\newblock \href {https://www.aclweb.org/anthology/C14-1022} {Sarcasm detection
  on {C}zech and {E}nglish {T}witter}.
\newblock In \emph{Proceedings of {COLING} 2014, the 25th International
  Conference on Computational Linguistics: Technical Papers}, pages 213--223,
  Dublin, Ireland. Dublin City University and Association for Computational
  Linguistics.

\bibitem[{Rajadesingan et~al.(2015)Rajadesingan, Zafarani, and
  Liu}]{rajadesingan2015sarcasm}
Ashwin Rajadesingan, Reza Zafarani, and Huan Liu. 2015.
\newblock \href {https://doi.org/10.1145/2684822.2685316} {Sarcasm detection on
  twitter: {A} behavioral modeling approach}.
\newblock In \emph{Proceedings of the Eighth {ACM} International Conference on
  Web Search and Data Mining, {WSDM} 2015, Shanghai, China, February 2-6,
  2015}, pages 97--106. {ACM}.

\bibitem[{Ren et~al.(2016)Ren, Zhang, Zhang, and Ji}]{ren2016context}
Yafeng Ren, Yue Zhang, Meishan Zhang, and Donghong Ji. 2016.
\newblock \href
  {http://www.aaai.org/ocs/index.php/AAAI/AAAI16/paper/view/11922}
  {Context-sensitive twitter sentiment classification using neural network}.
\newblock In \emph{Proceedings of the Thirtieth {AAAI} Conference on Artificial
  Intelligence, February 12-17, 2016, Phoenix, Arizona, {USA}}, pages 215--221.
  {AAAI} Press.

\bibitem[{Riloff et~al.(2013)Riloff, Qadir, Surve, De~Silva, Gilbert, and
  Huang}]{riloff2013sarcasm}
Ellen Riloff, Ashequl Qadir, Prafulla Surve, Lalindra De~Silva, Nathan Gilbert,
  and Ruihong Huang. 2013.
\newblock \href {https://www.aclweb.org/anthology/D13-1066} {Sarcasm as
  contrast between a positive sentiment and negative situation}.
\newblock In \emph{Proceedings of the 2013 Conference on Empirical Methods in
  Natural Language Processing}, pages 704--714, Seattle, Washington, USA.
  Association for Computational Linguistics.

\bibitem[{Rosenthal et~al.(2017)Rosenthal, Farra, and
  Nakov}]{rosenthal-2017-semeval}
Sara Rosenthal, Noura Farra, and Preslav Nakov. 2017.
\newblock \href {https://doi.org/10.18653/v1/S17-2088} {{S}em{E}val-2017 task
  4: Sentiment analysis in {T}witter}.
\newblock In \emph{Proceedings of the 11th International Workshop on Semantic
  Evaluation ({S}em{E}val-2017)}, pages 502--518, Vancouver, Canada.
  Association for Computational Linguistics.

\bibitem[{Sintsova and Pu(2016)}]{sintsova-2016-dystemo}
Valentina Sintsova and Pearl Pu. 2016.
\newblock \href {https://doi.org/10.1145/2912147} {Dystemo: Distant supervision
  method for multi-category emotion recognition in tweets}.
\newblock \emph{ACM Trans. Intell. Syst. Technol.}, 8(1).

\bibitem[{Song et~al.(2019)Song, Tan, Qin, Lu, and Liu}]{song2019mass}
Kaitao Song, Xu~Tan, Tao Qin, Jianfeng Lu, and Tie{-}Yan Liu. 2019.
\newblock \href {http://proceedings.mlr.press/v97/song19d.html} {{MASS:} masked
  sequence to sequence pre-training for language generation}.
\newblock In \emph{Proceedings of the 36th International Conference on Machine
  Learning, {ICML} 2019, 9-15 June 2019, Long Beach, California, {USA}},
  volume~97 of \emph{Proceedings of Machine Learning Research}, pages
  5926--5936. {PMLR}.

\bibitem[{Sun et~al.(2019)Sun, Wang, Li, Feng, Chen, Zhang, Tian, Zhu, Tian,
  and Wu}]{sun2019ernie}
Yu~Sun, Shuohuan Wang, Yu{-}Kun Li, Shikun Feng, Xuyi Chen, Han Zhang, Xin
  Tian, Danxiang Zhu, Hao Tian, and Hua Wu. 2019.
\newblock \href {http://arxiv.org/abs/1904.09223} {{ERNIE:} enhanced
  representation through knowledge integration}.
\newblock \emph{CoRR}, abs/1904.09223.

\bibitem[{Thomas(2014)}]{thomas2014meaning}
Jenny~A Thomas. 2014.
\newblock \emph{Meaning in interaction: An introduction to pragmatics}.
\newblock Routledge.

\bibitem[{Tian et~al.(2020)Tian, Gao, Xiao, Liu, He, Wu, Wang, and
  Wu}]{tian-2020-skep}
Hao Tian, Can Gao, Xinyan Xiao, Hao Liu, Bolei He, Hua Wu, Haifeng Wang, and
  Feng Wu. 2020.
\newblock \href {https://doi.org/10.18653/v1/2020.acl-main.374} {{SKEP}:
  Sentiment knowledge enhanced pre-training for sentiment analysis}.
\newblock In \emph{Proceedings of the 58th Annual Meeting of the Association
  for Computational Linguistics}, pages 4067--4076, Online. Association for
  Computational Linguistics.

\bibitem[{Van~der Maaten and Hinton(2008)}]{van2008visualizing}
Laurens Van~der Maaten and Geoffrey Hinton. 2008.
\newblock Visualizing data using t-sne.
\newblock \emph{Journal of machine learning research}, 9(11).

\bibitem[{Van~Hee et~al.(2018)Van~Hee, Lefever, and Hoste}]{van-hee2018semeval}
Cynthia Van~Hee, Els Lefever, and V{\'e}ronique Hoste. 2018.
\newblock \href {https://doi.org/10.18653/v1/S18-1005} {{S}em{E}val-2018 task
  3: Irony detection in {E}nglish tweets}.
\newblock In \emph{Proceedings of The 12th International Workshop on Semantic
  Evaluation}, pages 39--50, New Orleans, Louisiana. Association for
  Computational Linguistics.

\bibitem[{Vaswani et~al.(2017)Vaswani, Shazeer, Parmar, Uszkoreit, Jones,
  Gomez, Kaiser, and Polosukhin}]{vaswani2017attention}
Ashish Vaswani, Noam Shazeer, Niki Parmar, Jakob Uszkoreit, Llion Jones,
  Aidan~N. Gomez, Lukasz Kaiser, and Illia Polosukhin. 2017.
\newblock \href
  {https://proceedings.neurips.cc/paper/2017/hash/3f5ee243547dee91fbd053c1c4a845aa-Abstract.html}
  {Attention is all you need}.
\newblock In \emph{Advances in Neural Information Processing Systems 30: Annual
  Conference on Neural Information Processing Systems 2017, December 4-9, 2017,
  Long Beach, CA, {USA}}, pages 5998--6008.

\bibitem[{Wang et~al.(2019)Wang, Hula, Xia, Pappagari, McCoy, Patel, Kim,
  Tenney, Huang, Yu, Jin, Chen, Van~Durme, Grave, Pavlick, and
  Bowman}]{wang-2019-tell}
Alex Wang, Jan Hula, Patrick Xia, Raghavendra Pappagari, R.~Thomas McCoy, Roma
  Patel, Najoung Kim, Ian Tenney, Yinghui Huang, Katherin Yu, Shuning Jin,
  Berlin Chen, Benjamin Van~Durme, Edouard Grave, Ellie Pavlick, and Samuel~R.
  Bowman. 2019.
\newblock \href {https://doi.org/10.18653/v1/P19-1439} {Can you tell me how to
  get past sesame street? sentence-level pretraining beyond language modeling}.
\newblock In \emph{Proceedings of the 57th Annual Meeting of the Association
  for Computational Linguistics}, pages 4465--4476, Florence, Italy.
  Association for Computational Linguistics.

\bibitem[{Wang and Liu(2021)}]{wang-2021-understanding}
Feng Wang and Huaping Liu. 2021.
\newblock \href
  {https://openaccess.thecvf.com/content/CVPR2021/html/Wang\_Understanding\_the\_Behaviour\_of\_Contrastive\_Loss\_CVPR\_2021\_paper.html}
  {Understanding the behaviour of contrastive loss}.
\newblock In \emph{{IEEE} Conference on Computer Vision and Pattern
  Recognition, {CVPR} 2021, virtual, June 19-25, 2021}, pages 2495--2504.
  Computer Vision Foundation / {IEEE}.

\bibitem[{Waseem and Hovy(2016)}]{waseem-2016-hateful}
Zeerak Waseem and Dirk Hovy. 2016.
\newblock \href {https://doi.org/10.18653/v1/N16-2013} {Hateful symbols or
  hateful people? predictive features for hate speech detection on {T}witter}.
\newblock In \emph{Proceedings of the {NAACL} Student Research Workshop}, pages
  88--93, San Diego, California. Association for Computational Linguistics.

\bibitem[{Wiegand and Ruppenhofer(2021)}]{wiegand-2021-exploiting}
Michael Wiegand and Josef Ruppenhofer. 2021.
\newblock \href {https://www.aclweb.org/anthology/2021.eacl-main.28}
  {Exploiting emojis for abusive language detection}.
\newblock In \emph{Proceedings of the 16th Conference of the European Chapter
  of the Association for Computational Linguistics: Main Volume}, pages
  369--380, Online. Association for Computational Linguistics.

\bibitem[{Wolf et~al.(2020)Wolf, Debut, Sanh, Chaumond, Delangue, Moi, Cistac,
  Rault, Louf, Funtowicz, Davison, Shleifer, von Platen, Ma, Jernite, Plu, Xu,
  Le~Scao, Gugger, Drame, Lhoest, and Rush}]{wolf-2020-transformers}
Thomas Wolf, Lysandre Debut, Victor Sanh, Julien Chaumond, Clement Delangue,
  Anthony Moi, Pierric Cistac, Tim Rault, Remi Louf, Morgan Funtowicz, Joe
  Davison, Sam Shleifer, Patrick von Platen, Clara Ma, Yacine Jernite, Julien
  Plu, Canwen Xu, Teven Le~Scao, Sylvain Gugger, Mariama Drame, Quentin Lhoest,
  and Alexander Rush. 2020.
\newblock \href {https://doi.org/10.18653/v1/2020.emnlp-demos.6} {Transformers:
  State-of-the-art natural language processing}.
\newblock In \emph{Proceedings of the 2020 Conference on Empirical Methods in
  Natural Language Processing: System Demonstrations}, pages 38--45, Online.
  Association for Computational Linguistics.

\bibitem[{Wood and Ruder(2016)}]{wood2016emoji}
Ian Wood and Sebastian Ruder. 2016.
\newblock Emoji as emotion tags for tweets.
\newblock In \emph{Proceedings of the Emotion and Sentiment Analysis Workshop
  LREC2016, Portoro{\v{z}}, Slovenia}, pages 76--79.

\bibitem[{Yang et~al.(2019)Yang, Dai, Yang, Carbonell, Salakhutdinov, and
  Le}]{yang2019xlnet}
Zhilin Yang, Zihang Dai, Yiming Yang, Jaime~G. Carbonell, Ruslan Salakhutdinov,
  and Quoc~V. Le. 2019.
\newblock \href
  {https://proceedings.neurips.cc/paper/2019/hash/dc6a7e655d7e5840e66733e9ee67cc69-Abstract.html}
  {{XLNet}: Generalized autoregressive pretraining for language understanding}.
\newblock In \emph{Advances in Neural Information Processing Systems 32: Annual
  Conference on Neural Information Processing Systems 2019, NeurIPS 2019,
  December 8-14, 2019, Vancouver, BC, Canada}, pages 5754--5764.

\bibitem[{Zampieri et~al.(2019{\natexlab{a}})Zampieri, Malmasi, Nakov,
  Rosenthal, Farra, and Kumar}]{zampieri-2019-predicting}
Marcos Zampieri, Shervin Malmasi, Preslav Nakov, Sara Rosenthal, Noura Farra,
  and Ritesh Kumar. 2019{\natexlab{a}}.
\newblock \href {https://doi.org/10.18653/v1/N19-1144} {Predicting the type and
  target of offensive posts in social media}.
\newblock In \emph{Proceedings of the 2019 Conference of the North {A}merican
  Chapter of the Association for Computational Linguistics: Human Language
  Technologies, Volume 1 (Long and Short Papers)}, pages 1415--1420,
  Minneapolis, Minnesota. Association for Computational Linguistics.

\bibitem[{Zampieri et~al.(2019{\natexlab{b}})Zampieri, Malmasi, Nakov,
  Rosenthal, Farra, and Kumar}]{zampieri-2019-semeval}
Marcos Zampieri, Shervin Malmasi, Preslav Nakov, Sara Rosenthal, Noura Farra,
  and Ritesh Kumar. 2019{\natexlab{b}}.
\newblock \href {https://doi.org/10.18653/v1/S19-2010} {{S}em{E}val-2019 task
  6: Identifying and categorizing offensive language in social media
  ({O}ffens{E}val)}.
\newblock In \emph{Proceedings of the 13th International Workshop on Semantic
  Evaluation}, pages 75--86, Minneapolis, Minnesota, USA. Association for
  Computational Linguistics.

\bibitem[{Zhang et~al.(2019)Zhang, Han, Liu, Jiang, Sun, and
  Liu}]{zhang-2019-ernie}
Zhengyan Zhang, Xu~Han, Zhiyuan Liu, Xin Jiang, Maosong Sun, and Qun Liu. 2019.
\newblock \href {https://doi.org/10.18653/v1/P19-1139} {{ERNIE}: Enhanced
  language representation with informative entities}.
\newblock In \emph{Proceedings of the 57th Annual Meeting of the Association
  for Computational Linguistics}, pages 1441--1451, Florence, Italy.
  Association for Computational Linguistics.

\bibitem[{Zhou et~al.(2020)Zhou, Zhang, Zhao, and Zhang}]{zhou-2020-limit}
Junru Zhou, Zhuosheng Zhang, Hai Zhao, and Shuailiang Zhang. 2020.
\newblock \href {https://doi.org/10.18653/v1/2020.findings-emnlp.399}
  {{LIMIT}-{BERT} : Linguistics informed multi-task {BERT}}.
\newblock In \emph{Findings of the Association for Computational Linguistics:
  EMNLP 2020}, pages 4450--4461, Online. Association for Computational
  Linguistics.

\end{thebibliography}

\appendix
\appendixpage

\numberwithin{figure}{section}
\numberwithin{table}{section}
\section{Hyper-parameters and Procedure}\label{subsec:models:hyperparameter}

\textbf{Pragmatic Masking.} For pragmatic masking, we use the Adam optimizer with a weight decay of $0.01$~\cite{loshchilov2018decoupled} and a peak learning rate of $5e-5$. The number of the epochs is five. 

\noindent \textbf{Surrogate Fine-Tuning.} For surrogate fine-tuning, we fine-tune RoBERTa on surrogate classification tasks with the same Adam optimizer but use a peak learning rate of $2e-5$. 

The pre-training and surrogate fine-tuning models are trained on eight Nvidia V$100$ GPUs ($32$G each). On average the running time is $24$ hours per epoch for PMLMs, $2.5$ hours per epoch for SFT models. All the models are implemented by Huggingface Transformers~\cite{wolf-2020-transformers}.

\noindent \textbf{Downstream Fine-Tuning.} We evaluate the further pre-trained models with pragmatic masking and surrogate fine-tuned models on the $15$ downstream tasks in Table~\ref{tab:gold_data}. We set maximal sequence length as $60$ for $13$ text classification tasks. For Crisis\textsubscript{Oltea} and Stance\textsubscript{Moham}, we append the topic term behind the post content, separate them by [SEP] token, and set maximal sequence length to $70$, especially. For all the tasks, we pass the hidden state of [CLS] token from the last Transformer-encoder layer through a non-linear layer to predict. Cross-Entropy calculates the training loss. We then use Adam with a weight decay of $0.01$ to optimize the model and fine-tune each task for $20$ epochs with early stop ($patience = 5$ epochs). We fine-tune the peak learning rate in a set of $\{1e-5, 5e-6\}$ and batch size in a set of $\{8, 32, 64\}$. We find the learning rate of $5e-6$ performs best across all the tasks. For the downstream tasks whose Train set is smaller than $15,000$ samples, the best mini-batch size is eight. The best batch size of other larger downstream tasks is $64$. For fine-tuning BERTweet, we use the hyperparameters identified in~\citet{nguyen-etal-2020-bertweet}, i.e., a fixed learning rate of $1e-5$ and a batch size of $32$. 


We use the same hyperparameters to run three times with random seeds for all downstream fine-tuning (unless otherwise indicated). All downstream task models are fine-tuned on four Nvidia V$100$ GPUs ($32$G each). At the end of each epoch, we evaluate the model on the Dev set and identify the model that achieved the highest performance on Dev as our best model. We then test the best model on the Test set. In order to compute the model's overall performance across $15$ tasks, we use same evaluation metric (i.e., macro $F_1$) for all tasks. We report the average Test macro $F_1$ of the best model over three runs. We also average the macro $F_1$ scores across $15$ tasks to present the model's overall performance. 

\section{Few-Shot Experiment}\label{sec:append_fewshot}
 Tables~\ref{tab:few_shot-roberta}, \ref{tab:few_shot-bertweet}, \ref{tab:few_shot-x1sfte}, and \ref{tab:few_shot-x2sfth} respectively, present the performance of RoBERTa, BERTweet, PragS1, and PragS2 on all our $15$ English downstream datasets and various few-shot settings.

\begin{table*}[ht]
\small
\centering
\begin{tabular}{@{}lccccccccccc@{}}
\toprule
\multicolumn{1}{c}{\textbf{Task}}                & \textbf{1} & \textbf{5} & \textbf{10} & \textbf{20} & \textbf{30} & \textbf{40} & \textbf{50} & \textbf{60} & \textbf{70} & \textbf{80} & \textbf{90} \\ \midrule
Crisis\textsubscript{Oltea-14}  & 94.67      & 95.36      & 95.55       & 95.74       & 95.90       & 95.81       & 95.89       & 95.84       & 95.99       & 96.03       & 96.11       \\
Emo\textsubscript{Moham-18}     & 14.10      & 30.36      & 71.76       & 73.62       & 76.26       & 77.02       & 77.59       & 77.19       & 77.38       & 77.84       & 78.86       \\
Hate\textsubscript{Waseem-16}   & 28.23      & 52.66      & 54.66       & 54.82       & 56.26       & 56.42       & 56.70       & 57.10       & 56.92       & 56.99       & 57.25       \\
Hate\textsubscript{David-17}    & 42.01      & 70.92      & 74.76       & 75.71       & 75.08       & 75.70       & 76.05       & 75.21       & 76.38       & 76.58       & 77.63       \\
Humor\textsubscript{Potash-17}  & 47.91      & 47.91      & 52.89       & 52.67       & 54.43       & 52.30       & 53.89       & 55.00       & 53.69       & 54.16       & 56.78       \\
Humor\textsubscript{Meaney-21}  & 53.44      & 89.50      & 89.47       & 90.12       & 91.95       & 91.65       & 92.33       & 91.96       & 92.65       & 91.78       & 92.27       \\
Irony\textsubscript{Hee-18A}    & 40.75      & 60.47      & 61.97       & 70.49       & 67.64       & 70.40       & 72.04       & 71.33       & 72.01       & 72.67       & 72.54       \\
Irony\textsubscript{Hee-18B}    & 19.41      & 26.27      & 43.61       & 46.47       & 44.78       & 48.41       & 50.40       & 51.65       & 51.80       & 53.15       & 53.17       \\
Offense\textsubscript{-Zamp-19} & 41.89      & 76.87      & 74.44       & 76.53       & 79.75       & 79.29       & 78.95       & 78.13       & 79.01       & 79.42       & 79.90       \\
Sarc\textsubscript{Riloff-13}   & 44.41      & 44.80      & 43.99       & 70.49       & 51.10       & 70.70       & 67.72       & 72.46       & 67.98       & 72.88       & 73.75       \\
Sarc\textsubscript{Ptacek-14}   & 81.57      & 85.92      & 87.18       & 88.78       & 89.84       & 91.33       & 91.76       & 92.38       & 93.58       & 94.29       & 94.98       \\
Sarc\textsubscript{Rajad-15}    & 68.52      & 77.80      & 78.47       & 81.59       & 82.60       & 82.58       & 83.61       & 83.77       & 84.44       & 84.76       & 84.43       \\
Sarc\textsubscript{Bam-15}      & 64.17      & 74.01      & 75.95       & 76.18       & 77.00       & 78.07       & 78.43       & 78.68       & 79.35       & 79.08       & 79.40       \\
Senti\textsubscript{Rosen-17}   & 64.84      & 68.00      & 69.95       & 70.10       & 70.51       & 70.04       & 71.70       & 70.07       & 70.12       & 70.30       & 71.17       \\
Stance\textsubscript{Moham-16}  & 25.20      & 44.73      & 62.03       & 62.67       & 65.11       & 65.44       & 64.97       & 65.74       & 68.59       & 68.54       & 69.21       \\ \cdashline{1-12}
Average                                          & 48.74      & 63.04      & 69.11       & 72.40       & 71.88       & 73.68       & 74.14       & 74.43       & 74.66       & 75.23       & 75.83       \\ \bottomrule
\end{tabular}
\caption{Full result of few-shot learning on Baseline (1), fine-tuning RoBERTa.}\label{tab:few_shot-roberta}
\end{table*}

\begin{table*}[ht]
\small
\centering
\begin{tabular}{@{}lccccccccccc@{}}
\toprule
\textbf{Task}                                    & \textbf{1} & \textbf{5} & \textbf{10} & \textbf{20} & \textbf{30} & \textbf{40} & \textbf{50} & \textbf{60} & \textbf{70} & \textbf{80} & \textbf{90} \\ \midrule
Crisis\textsubscript{Oltea-14}  & 94.71      & 94.95      & 95.38       & 95.32       & 95.60       & 95.53       & 95.78       & 95.72       & 95.65       & 95.71       & 95.68       \\
Emo\textsubscript{Moham-18}     & 21.68      & 17.29      & 66.13       & 75.03       & 76.50       & 77.72       & 76.20       & 79.16       & 79.22       & 79.37       & 80.58       \\
Hate\textsubscript{Waseem-16}   & 30.92      & 52.27      & 53.70       & 55.05       & 55.18       & 55.80       & 56.48       & 56.44       & 56.46       & 57.10       & 56.66       \\
Hate\textsubscript{David-17}    & 29.21      & 69.18      & 74.17       & 76.58       & 77.95       & 76.97       & 77.19       & 77.43       & 77.29       & 77.72       & 78.30       \\
Humor\textsubscript{Potash-17}  & 47.90      & 47.91      & 48.24       & 51.68       & 51.25       & 53.37       & 54.80       & 54.39       & 54.91       & 52.31       & 55.83       \\
Humor\textsubscript{Meaney-21}  & 52.07      & 90.67      & 92.43       & 92.68       & 93.50       & 93.32       & 92.88       & 93.52       & 94.31       & 94.18       & 94.55       \\
Irony\textsubscript{Hee-18A}    & 44.88      & 57.78      & 67.90       & 71.87       & 74.40       & 75.42       & 75.15       & 75.94       & 75.42       & 76.80       & 76.82       \\
Irony\textsubscript{Hee-18B}    & 17.16      & 20.69      & 27.30       & 39.72       & 46.40       & 49.26       & 50.29       & 51.41       & 54.08       & 54.08       & 55.49       \\
Offense\textsubscript{-Zamp-19} & 45.03      & 74.68      & 76.49       & 78.02       & 79.26       & 78.55       & 78.86       & 79.59       & 80.54       & 79.74       & 78.30       \\
Sarc\textsubscript{Riloff-13}   & 44.38      & 43.99      & 44.88       & 43.99       & 77.89       & 78.23       & 77.73       & 79.73       & 78.20       & 79.98       & 78.82       \\
Sarc\textsubscript{Ptacek-14}   & 85.36      & 88.06      & 89.18       & 90.58       & 91.44       & 92.60       & 93.44       & 93.64       & 94.40       & 95.30       & 95.77       \\
Sarc\textsubscript{Rajad-15}    & 47.01      & 81.87      & 83.24       & 84.22       & 85.31       & 85.38       & 85.73       & 85.86       & 86.11       & 86.77       & 86.76       \\
Sarc\textsubscript{Bam-15}      & 56.24      & 76.75      & 78.61       & 80.01       & 80.06       & 81.05       & 81.05       & 81.64       & 81.86       & 82.72       & 82.84       \\
Senti\textsubscript{Rosen-17}   & 65.42      & 67.96      & 69.85       & 70.38       & 71.24       & 71.49       & 71.76       & 71.29       & 71.49       & 72.29       & 71.63       \\
Stance\textsubscript{Moham-16}  & 25.69      & 25.36      & 24.27       & 59.25       & 61.58       & 63.45       & 62.31       & 65.08       & 66.64       & 66.54       & 67.63       \\ \cdashline{1-12}
Average                                          & 47.18      & 60.63      & 66.12       & 70.96       & 74.50       & 75.21       & 75.31       & 76.06       & 76.44       & 76.71       & 77.04       \\ \bottomrule
\end{tabular}
\caption{Full result of few-shot learning on BERTweet.}\label{tab:few_shot-bertweet}
\end{table*}

\begin{table*}[ht]
\small
\centering
\begin{tabular}{@{}lccccccccccc@{}}
\toprule
\multicolumn{1}{c}{\textbf{Task}}                & \textbf{1} & \textbf{5} & \textbf{10} & \textbf{20} & \textbf{30} & \textbf{40} & \textbf{50} & \textbf{60} & \textbf{70} & \textbf{80} & \textbf{90} \\ \midrule
Crisis\textsubscript{Oltea-14}  & 94.35      & 95.34      & 95.37       & 95.74       & 95.85       & 95.83       & 95.92       & 95.92       & 95.91       & 95.98       & 95.86       \\
Emo\textsubscript{Moham-18}     & 36.95      & 64.31      & 74.68       & 77.94       & 79.79       & 80.19       & 80.23       & 80.19       & 80.30       & 80.78       & 81.27       \\
Hate\textsubscript{Waseem-16}   & 38.81      & 51.76      & 53.54       & 54.32       & 55.70       & 56.00       & 56.49       & 56.43       & 57.06       & 59.56       & 59.76       \\
Hate\textsubscript{David-17}    & 57.07      & 68.95      & 72.66       & 75.03       & 75.14       & 75.11       & 75.86       & 77.53       & 77.09       & 76.11       & 76.88       \\
Humor\textsubscript{Potash-17}  & 47.91      & 50.24      & 51.87       & 51.21       & 51.92       & 54.91       & 53.26       & 52.22       & 52.37       & 54.36       & 54.39       \\
Humor\textsubscript{Meaney-21}  & 87.10      & 91.79      & 92.16       & 92.42       & 92.80       & 93.01       & 93.05       & 93.53       & 93.64       & 93.86       & 93.70       \\
Irony\textsubscript{Hee-18A}    & 60.35      & 66.13      & 70.77       & 72.26       & 74.24       & 73.82       & 74.95       & 74.92       & 75.97       & 75.87       & 77.37       \\
Irony\textsubscript{Hee-18B}    & 29.82      & 36.42      & 41.72       & 46.50       & 50.14       & 53.57       & 52.63       & 55.80       & 54.23       & 55.92       & 56.62       \\
Offense\textsubscript{-Zamp-19} & 61.17      & 74.22      & 77.05       & 77.63       & 79.22       & 80.62       & 79.09       & 80.77       & 81.27       & 79.85       & 79.68       \\
Sarc\textsubscript{Riloff-13}   & 52.83      & 63.39      & 73.40       & 74.34       & 77.10       & 78.01       & 77.87       & 77.53       & 77.32       & 77.32       & 78.72       \\
Sarc\textsubscript{Ptacek-14}   & 85.64      & 87.81      & 88.87       & 89.90       & 91.17       & 92.18       & 92.82       & 93.64       & 94.00       & 95.08       & 95.68       \\
Sarc\textsubscript{Rajad-15}    & 82.80      & 84.95      & 85.84       & 85.79       & 86.62       & 86.39       & 86.84       & 86.96       & 86.81       & 87.14       & 87.02       \\
Sarc\textsubscript{Bam-15}      & 72.44      & 77.74      & 78.97       & 80.27       & 81.08       & 81.74       & 81.56       & 81.62       & 81.98       & 81.53       & 82.29       \\
Senti\textsubscript{Rosen-17}   & 59.48      & 65.39      & 69.06       & 69.29       & 70.18       & 70.32       & 71.51       & 71.42       & 71.28       & 71.87       & 72.13       \\
Stance\textsubscript{Moham-16}  & 31.80      & 49.63      & 56.29       & 60.94       & 64.59       & 64.58       & 65.44       & 67.27       & 68.23       & 67.95       & 68.13       \\ \cdashline{1-12}
Average                                          & 59.90      & 68.54      & 72.15       & 73.57       & 75.04       & 75.75       & 75.83       & 76.38       & 76.50       & 76.88       & 77.30       \\ \bottomrule
\end{tabular}
\caption{Full result of few-shot learning on PragS1.}\label{tab:few_shot-x1sfte}
\end{table*}

\begin{table*}[ht]
\small
\centering
\begin{tabular}{@{}lccccccccccc@{}}
\toprule
\multicolumn{1}{c}{\textbf{Task}}                & \textbf{1} & \textbf{5} & \textbf{10} & \textbf{20} & \textbf{30} & \textbf{40} & \textbf{50} & \textbf{60} & \textbf{70} & \textbf{80} & \textbf{90} \\ \midrule
Crisis\textsubscript{Oltea-14}  & 93.92      & 95.07      & 95.50       & 95.30       & 95.60       & 95.50       & 95.73       & 95.66       & 95.52       & 95.70       & 95.96       \\
Emo\textsubscript{Moham-18}     & 35.90      & 58.23      & 71.27       & 75.36       & 77.71       & 78.80       & 79.25       & 78.99       & 79.74       & 80.06       & 81.28       \\
Hate\textsubscript{Waseem-16}   & 43.42      & 53.24      & 59.36       & 54.85       & 55.51       & 56.32       & 56.57       & 56.52       & 56.91       & 61.08       & 63.86       \\
Hate\textsubscript{David-17}    & 57.30      & 71.09      & 73.10       & 75.37       & 77.25       & 74.36       & 75.91       & 77.72       & 75.76       & 77.30       & 76.59       \\
Humor\textsubscript{Potash-17}  & 49.75      & 51.72      & 51.59       & 52.39       & 54.80       & 53.39       & 52.82       & 52.31       & 53.41       & 53.82       & 54.26       \\
Humor\textsubscript{Meaney-21}  & 84.95      & 92.09      & 92.73       & 93.16       & 94.17       & 94.07       & 93.54       & 93.57       & 93.81       & 93.52       & 93.89       \\
Irony\textsubscript{Hee-18A}    & 57.95      & 68.51      & 71.96       & 73.41       & 75.17       & 75.66       & 75.60       & 77.34       & 76.72       & 77.49       & 77.79       \\
Irony\textsubscript{Hee-18B}    & 29.69      & 35.93      & 41.51       & 48.44       & 52.77       & 52.71       & 55.87       & 56.07       & 58.13       & 55.63       & 55.43       \\
Offense\textsubscript{-Zamp-19} & 52.61      & 70.40      & 74.09       & 76.45       & 78.80       & 78.02       & 76.90       & 79.53       & 79.35       & 79.73       & 79.42       \\
Sarc\textsubscript{Riloff-13}   & 49.57      & 64.07      & 75.80       & 75.46       & 78.28       & 78.93       & 78.89       & 78.31       & 79.71       & 78.86       & 79.04       \\
Sarc\textsubscript{Ptacek-14}   & 86.19      & 88.52      & 89.53       & 90.75       & 91.55       & 92.21       & 93.03       & 93.73       & 94.28       & 95.04       & 95.71       \\
Sarc\textsubscript{Rajad-15}    & 84.69      & 85.43      & 85.61       & 86.48       & 87.13       & 86.86       & 87.08       & 87.05       & 87.36       & 87.29       & 87.48       \\
Sarc\textsubscript{Bam-15}      & 73.40      & 77.28      & 77.88       & 79.84       & 79.40       & 80.29       & 80.31       & 80.32       & 80.60       & 80.95       & 80.39       \\
Senti\textsubscript{Rosen-17}   & 55.75      & 62.50      & 66.50       & 68.90       & 70.09       & 70.64       & 70.89       & 71.32       & 71.34       & 71.51       & 71.64       \\
Stance\textsubscript{Moham-16}  & 34.36      & 47.62      & 56.00       & 61.47       & 63.45       & 66.13       & 65.47       & 67.09       & 68.60       & 68.09       & 69.06       \\ \cdashline{1-12}
Average                                          & 59.30      & 68.11      & 72.16       & 73.84       & 75.44       & 75.59       & 75.86       & 76.37       & 76.75       & 77.07       & 77.45       \\ \bottomrule
\end{tabular}
\caption{Full result of few-shot learning on PragS2.}\label{tab:few_shot-x2sfth}
\end{table*}
\section{Uniformity and Tolerance}\label{sec:uni-tole}
\citet{wang-2021-understanding} investigate representation quality measuring the uniformity of an embedding distribution and the tolerance to semantically similar samples. Given a dataset $D$ and an encoder $\Phi$, the uniformity metric is based on a gaussian potential kernel and is formulated as:
\begin{equation}
    Uniformity = log \mathop{\mathbb{E}}_{x_i, x_j\in D}[ e^{ t ||\Phi(x_i) - \Phi(x_j)||^2_2 } ], 
\end{equation}
where $t=2$.

The tolerance metric measures the mean of similarities of samples belonging to the same class, which defined as:
\begin{equation}
\small
    Tolerance = \mathop{\mathbb{E}}_{x_i, x_j\in D}[(\Phi(x_i)^T\Phi(x_j)) \cdot I_{l(x_i)=l(x_j)}],
\end{equation}

where $l(x_i)$ is the supervised label of sample $x_i$. $I_{l(x_i)=l(x_j)}$ is an indicator function, giving the value of $1$ for $l(x_i)=l(x_j)$ and the value of $0$ for $l(x_i)\neq l(x_j)$. In our experiments, we use gold development samples from $13$ our social meaning datasets. 



\end{document}